\documentclass[11pt]{article}

\usepackage[final]{acl}

\usepackage{times}
\usepackage{latexsym}

\usepackage[T1]{fontenc}

\usepackage[utf8]{inputenc}

\usepackage{microtype}

\usepackage{inconsolata}

\usepackage{graphicx}

\usepackage[utf8]{inputenc} 
\usepackage[T1]{fontenc}    
\usepackage{url}            
\usepackage{booktabs}       
\usepackage{amsfonts}       
\usepackage{nicefrac}       
\usepackage{microtype}      
\usepackage{xcolor}
\definecolor{bluecite}{HTML}{0071BC}

\usepackage[toc]{appendix}
\usepackage{etoc}
\usepackage{minitoc}
\etocsettocstyle{\section*{Contents of Technical Appendices}}{}

\usepackage{graphicx}
\usepackage{subcaption}
\usepackage{multirow}

\usepackage{amsmath}

\usepackage[capitalize]{cleveref}
\crefname{section}{Sec.}{Secs.}
\Crefname{section}{Section}{Sections}
\Crefname{table}{Table}{Tables}
\crefname{table}{Tab.}{Tabs.}

\usepackage{microtype}
\usepackage{booktabs} %

\definecolor{darkgreen}{rgb}{0.0, 0.5, 0.0} 

\newcommand{\comt}[1]{#1}

\renewcommand{\comt}[1]{}

\usepackage[dvipsnames,table]{xcolor}

\usepackage{tabularx}
\usepackage{multicol}
\definecolor{myblue}{RGB}{235,235,250}

\usepackage{xspace}

\usepackage{pifont}
\usepackage{latexsym}
\usepackage{inconsolata}
\usepackage{arydshln}
\usepackage{enumitem}
\usepackage{wrapfig}
\usepackage{array}
\usepackage{epigraph}

\definecolor{lightpink}{RGB}{204, 231, 207} 
\definecolor{lightblue}{RGB}{210, 220, 250} 
\definecolor{lightgray}{RGB}{237, 237, 237} 

\definecolor{superlightred}{rgb}{0.99, 0.92, 0.92}
\definecolor{darkgreen}{RGB}{50,100,0}
\definecolor{darkred}{RGB}{200, 0, 0}

\usepackage{makecell}
\usepackage{colortbl}

\usepackage{hyperref}
\usepackage{url}
\usepackage{hyperref}
\usepackage[most]{tcolorbox}
\usepackage{wrapfig}
\usepackage{graphicx,xcolor,float}
\usepackage{subcaption}
\usepackage{threeparttable}
\usepackage{algorithm}
\usepackage{pifont}
\usepackage{colortbl}
\usepackage{color}
\usepackage{multirow}
\usepackage{tabularx}
\usepackage{float}
\usepackage{graphicx}
\usepackage{booktabs}
\usepackage{arydshln}
\usepackage{enumitem}
\usepackage{wrapfig}
\usepackage{caption}
\usepackage{graphicx}
\usepackage{makecell}
\usepackage{float} 
\usepackage{makecell}
\usepackage{tabularx}
\usepackage{twemojis}
\usepackage{amssymb,mathrsfs,amsmath}
\usepackage{pifont}
\usepackage[export]{adjustbox}
\usepackage{xcolor}
\usepackage{xspace}
\usepackage{shadowtext}
\usepackage{anyfontsize}

\usepackage{threeparttable}

\hypersetup{
    colorlinks=true,
    linkcolor=red,
    citecolor=cyan,
    filecolor=magenta,      
    urlcolor=magenta,
    }

\definecolor{bittersweet}{rgb}{1.0, 0.44, 0.37}
\definecolor{mygreen}{rgb}{0.29, 0.7, 0.48}
\definecolor{my_green}{RGB}{51,102,0}
\definecolor{my_yellow}{RGB}{255,165,0}
\definecolor{my_red}{RGB}{204, 0, 0}

\usepackage{pifont}
\definecolor{mygray}{gray}{0.4}
\hypersetup{
    colorlinks=true,
    linkcolor=red,
    citecolor=cyan,
    filecolor=magenta,      
    urlcolor=magenta,
    }
\usepackage{xcolor} 
\usepackage{xcolor}

\usepackage[font=small,labelfont=bf]{caption}  
\usepackage{makecell}
\usepackage{tabulary}
\definecolor{ada_green}{rgb}{0,205,205}
\definecolor{glt_red}{rgb}{109,205,255}

\definecolor{backred}{RGB}{255, 190, 190}
\definecolor{backblue}{RGB}{210, 230, 250}
\definecolor{backgrey}{RGB}{220, 220, 220}
\newcommand{\high}{\cellcolor{backblue}}

\definecolor{shadecolor}{RGB}{237,237,237}

\usepackage[utf8]{inputenc} 
\usepackage[T1]{fontenc}    
\usepackage{url}            
\usepackage{booktabs}       
\usepackage{amsfonts}       
\usepackage{nicefrac}       
\usepackage{microtype}      
\usepackage[table]{xcolor}
\definecolor{bluecite}{HTML}{0071BC}

\usepackage[toc]{appendix}
\usepackage{etoc}
\usepackage{minitoc}
\usepackage{makecell}

\title{Position: Multimodal Large Language Models Can Significantly \\Advance Scientific Reasoning}

\author{\textbf{Yibo Yan}$^{1,2,3}$, 
    \textbf{Shen Wang}$^{1}$, 
    \textbf{Jiahao Huo}$^{2}$, 
    \textbf{Jingheng Ye}$^{1,4}$, 
    \textbf{Zhendong Chu}$^{1}$,\\
    \textbf{Xuming Hu}$^{2,3,}$\thanks{Corresponding Authors} ,
    \textbf{Philip S. Yu}$^{5}$,
    \textbf{Carla Gomes}$^{6}$,
    \textbf{Bart Selman}$^{6}$,
    \textbf{Qingsong Wen}$^{1,}$\footnotemark[1]\\
    $^1$Squirrel AI,
    $^2$HKUST(GZ), 
    $^3$HKUST, 
    $^4$Tsinghua University,\\
    $^5$University of Illinois at Chicago,
    $^6$Cornell University\\
    \texttt{\href{mailto:yanyibo70@gmail.com}{yanyibo70@gmail.com}},
     \texttt{\href{mailto:xuminghu@hkust-gz.edu.cn}{xuminghu@hkust-gz.edu.cn}},
     \texttt{\href{mailto:qingsongedu@gmail.com}{qingsongedu@gmail.com}}
    \\
}

\begin{document}
\maketitle

\etocdepthtag.toc{mtchapter}
\etocsettagdepth{mtchapter}{subsection}
\etocsettagdepth{mtappendix}{none}

\begin{abstract}
Scientific reasoning, the process through which humans apply logic, evidence, and critical thinking to explore and interpret scientific phenomena, is essential in advancing knowledge reasoning across diverse fields. However, despite significant progress, current scientific reasoning models still \textit{struggle with generalization across domains and often fall short of multimodal perception}. Multimodal Large Language Models (MLLMs), which integrate text, images, and other modalities, present an exciting opportunity to overcome these limitations and enhance scientific reasoning. Therefore, \textbf{this position paper argues that MLLMs can significantly advance scientific reasoning} across disciplines such as mathematics, physics, chemistry, and biology. First, we propose a four-stage research roadmap of scientific reasoning capabilities, and highlight the current state of MLLM applications in scientific reasoning, noting their ability to integrate and reason over diverse data types. Second, we summarize the key challenges that remain obstacles to achieving MLLM's full potential. To address these challenges, we propose actionable insights and suggestions for the future. Overall, our work offers a novel perspective on MLLM integration with scientific reasoning, providing the LLM community with a valuable vision for achieving Artificial General Intelligence (AGI).
\end{abstract}

\section{Introduction}
\label{sec:intro}

Scientific reasoning, at its core, is the process through which humans apply logic, evidence, and critical thinking to explore and interpret phenomena in various scientific domains \cite{bao2009learning,lawson2004nature}. This cognitive ability is essential not only for advancing knowledge but also for fostering a deeper understanding of the natural world, particularly in fields such as mathematics, physics, chemistry, and biology. In education, medicine, finance, AI for Science, and other domains, scientific reasoning serves as a cornerstone for cultivating problem-solving skills, analytical thinking, and innovation \cite{jadon2025enhancing}. However, despite shared objectives, each domain has unique characteristics in terms of data representation, knowledge construction, and reasoning methods \cite{ferrag2025llm,chen2025towards}.

\begin{figure*}[htbp]
    \centering
    \includegraphics[width=0.9\linewidth]{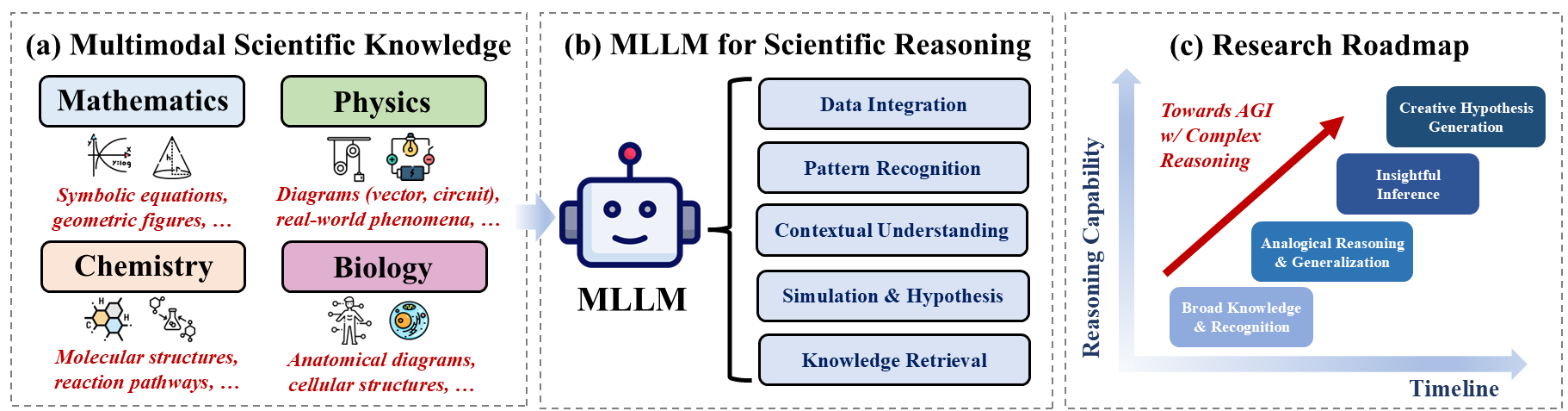}
    \vspace{-2mm}
    \caption{The big picture of our position. We focus on multimodal scientific fields,  especially mathematics, physics, chemistry, and biology as our scope (a), and we advocate leveraging MLLMs with multiple reasoning functions for scientific reasoning (b). We further propose a four-stage roadmap for scientific reasoning capability, ultimately achieving AGI (c).}
    \label{fig:roadmap}
    \vspace{-4mm}
\end{figure*}

In response to these challenges, the scientific community has explored a range of approaches, from traditional statistical methods to the more recent advancements in deep learning, with the goal of improving knowledge reasoning across disciplines \cite{goodman2016aligning,lu2022survey}. While significant progress has been made in enhancing scientific reasoning within specific domains, a gap remains in the broader context of scientific research. Current scientific reasoning models, and even those targeted toward domain-specific applications, are still \textit{far from achieving the generalization capabilities necessary for Artificial General Intelligence (AGI), which aims to exhibit unified reasoning across all fields} \cite{birhane2023science}.

\begin{figure}[t]
    \centering
    \includegraphics[width=0.7\linewidth]{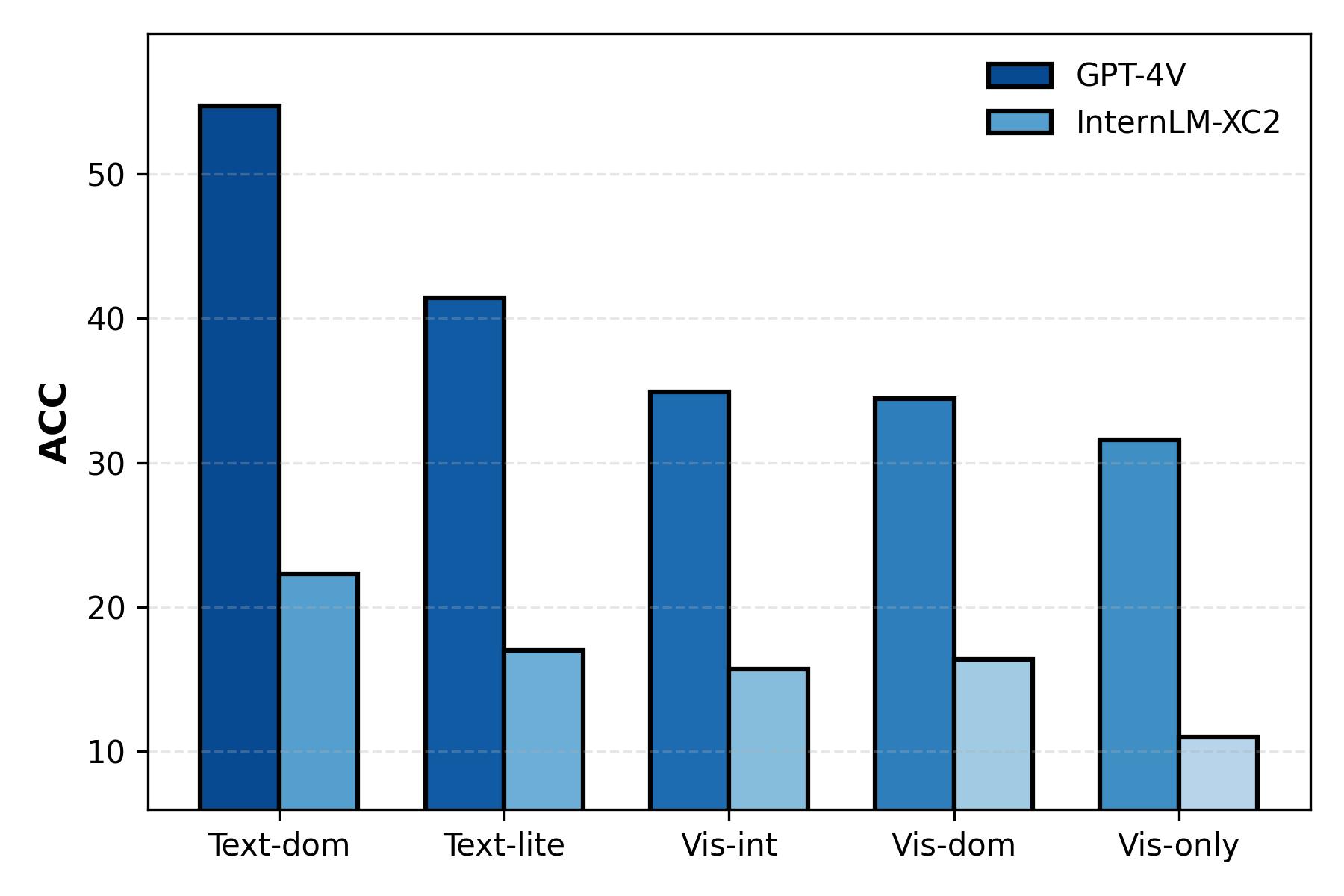}
    \vspace{-2mm}
    \caption{MLLM performance declines w/ increased reliance on visual modality from MathVerse \cite{zhang2025mathverse}.}
    \label{fig:multimodal_components}
    \vspace{-6mm}
\end{figure}

The rapid rise of Large Language Models (LLMs) in recent years has brought transformative changes across various domains, pushing the boundaries of what is possible in natural language processing and understanding \cite{min2023recent,zhao2023survey}. Despite their remarkable zero-shot reasoning abilities, many areas, particularly in scientific fields, require multimodal inputs to build a comprehensive understanding of knowledge. This has led to the emergence and growth of Multimodal Large Language Models (MLLMs), which are capable of integrating and reasoning over multiple types of data, such as text, images, and other modalities \cite{liang2024survey,bai2024survey}. MLLMs are not only revolutionizing language understanding but also paving the way for advancements in scientific reasoning by processing complex multimodal data in ways that were previously unachievable. However, a critical gap in visual reasoning capabilities persists, as MLLM performance degrades when shifting reliance from textual descriptions to visual diagrams (Figure \ref{fig:multimodal_components}).

Building on the rapid evolution of MLLMs and the growing demand for enhanced multimodal reasoning capabilities in scientific domains \cite{reddy2024towards,zhang2024scientific}, \textbf{this paper proposes a position: MLLMs can significantly advance scientific reasoning}. By integrating multimodal learning techniques, MLLMs have the potential to address the pressing challenges in scientific reasoning, as shown in Figure \ref{fig:roadmap}. This paper aims to break down the analysis into three key sections: the current state of MLLMs in scientific reasoning, the challenges encountered, and the future steps needed to achieve greater success.

Our analysis delves into these topics in detail, starting with an overview of background and research roadmap (Section \ref{sec:background}), followed by how MLLMs are currently handling scientific reasoning (Section \ref{sec:how_reason_now}) as well as a discussion of the challenges faced (Section \ref{sec:challenges}), and concluding with the promising opportunities that lie ahead (Section \ref{sec:discussion}). In particular, we first explore how MLLMs have been applied across scientific disciplines, detailing the techniques used for data integration, pattern recognition, contextual understanding, and more. The Challenges section examines the inherent difficulties, ranging from technical to ethical aspects, that need to be overcome for MLLMs to achieve optimal performance. Finally, in the Discussion section, we highlight future steps with relevant literature, from data \& training strategies to agent-based collaboration, as key areas for advancing the integration of MLLMs with scientific reasoning. Through this, we aim to provide a comprehensive understanding of the landscape and offer strategic insights for future research in the intersection of MLLMs and scientific reasoning\footnote{See more clarification of our scope in Appendix \ref{app:scope_clarification}.}.

Our position and analytical framework contribute in three significant ways. \ding{182} We present \textbf{a novel perspective} on the integration of MLLMs with scientific reasoning, highlighting how this synthesis could reshape research in the field. \ding{183} We offer \textbf{a systematic review and categorization} of recent advancements, showcasing representative work and outlining a clear roadmap for future progress. \ding{184} We identify and explore \textbf{future opportunities}, providing the community with promising insights that could guide the next generation of research in scientific reasoning.
\section{Background}
\label{sec:background}

\subsection{Scientific Reasoning}
\label{sec:sci_reasoning_definition}
Scientific reasoning is the intellectual process of forming hypotheses, interpreting evidence, and applying logical frameworks to solve problems or explain phenomena  \cite{bao2009learning,lawson2004nature}. Its importance spans diverse scientific domains, such as mathematics, physics, chemistry, and biology, where it drives discovery, fosters understanding, and enables practical innovation. With the rise of multimodal data, scientific reasoning increasingly requires integrating and synthesizing information from multiple sources, including textual, visual, and other modalities\footnote{See the formal formulation of the task in Appendix \ref{app:task_formulation}.}.

The significance of scientific reasoning in the MLLM context is profound. By enabling models to connect disparate data points and infer relationships across modalities, MLLMs hold the potential to transform how researchers approach interdisciplinary problems. This capability is critical for addressing grand challenges such as climate modeling, drug discovery, \textit{etc} \cite{zhang2024scientific}. Moreover, enhancing MLLM-based scientific reasoning aligns with broader goal of advancing AGI, as it exemplifies synthesis of learning, abstraction, and decision-making across domains \cite{Guo2025SciVerseUT,thawakar2025llamavo1}.

\subsection{Multimodal Large Language Models}
\label{sec:mllm_definition}
Most existing MLLMs consist of three primary modules: a modality encoder, an LLM module, and a projector between them \cite{fu2024mme}. Typically, the modality encoder extracts embeddings from non-language modalities such as images or audio, which are then projected into the word space of the LLM via the projector. The post-projection embeddings are subsequently combined with word embeddings derived from system prompts and user queries to serve as input for the LLM. Similar to LLMs, MLLMs generate responses in an autoregressive manner:$$p(\mathbf{w}_O \mid \mathbf{w}_V, \mathbf{w}_T) \sim \prod_{t=1}^L P(w_t \mid w_{<t}, \mathbf{w}_V, \mathbf{w}_T)$$ Here, $\mathbf{w}_V$ and $\mathbf{w}_T$ denote the post-projection embeddings and word embeddings respectively, while $\mathbf{w}_O = \{w_{o,t}\}_{t=1}^L$ represents the generated word token sequence of length $L$. With their capability to comprehend visual inputs, contemporary MLLMs demonstrate remarkable performance in various tasks, including visual question answering (VQA)~\cite{ishmam2024image,uppal2022multimodal,dang2024exploring}, image captioning~\cite{vaishnavi2024video,agarwal2024methods}, and multimodal reasoning~\cite{yan2024georeasoner,yan2024survey,huo2024mmneuron}.
\par Presently, there is a plethora of open-source foundation MLLMs capable of general multimodal tasks. Notable examples include LLaVA family~\cite{liu2023llava,liu2023improvedllava,liu2024llavanext}, Qwen-VL series~\cite{Qwen-VL, Qwen2VL}, InternVL series~\cite{chen2024internvl,chen2024far}, LLaMA-3.2-Vision~\cite{dubey2024llama3}, etc. Despite these advancements, open-source MLLMs still lag behind closed-source models like GPT-4o~\cite{achiam2023gpt4o}, Claude~\cite{anthropic2024claude3.5}, and Gemini-Pro~\cite{team2024gemini1.5} in complex reasoning tasks~\cite{liu2025mmbench,yue2024mmmu}. With the emergence of o1-like reasoning models~\cite{jaech2024o1,zeng2025revisiting,liu2025comprehensive}, preliminary efforts are underway to elicit the slow-thinking capabilities of MLLMs, as seen in works like QvQ~\cite{qvq-72b-preview}, Mulberry~\cite{yao2024mulberry}, Virgo \cite{du2025virgo}, \textit{etc}.
\section{How MLLMs Benefit Scientific Reasoning}
\label{sec:how_reason_now}

\subsection{Research Roadmap}
\label{sec:roadmap}
The development of (M)LLMs for scientific reasoning can be categorized into four progressive stages: \textit{Broad Knowledge and Recognition}, \textit{Analogical Reasoning and Generalization}, \textit{Insightful Inference}, and \textit{Creative Hypothesis Generation}. Each stage is defined by its unique characteristics across four dimensions: data and knowledge requirements, reasoning mechanisms, model generalization, and applications and impact (See the detailed comparisons among the four dimensions in Appendix \ref{app:roadmap}). Figure \ref{fig:roadmap} (c) provides an overview of four stages, highlighting their evolution progress.

\textbf{Stage 1: Broad Knowledge and Recognition.} The initial stage focuses on building a strong foundational understanding across domains. MLLMs in this stage rely on highly diverse and multimodal datasets to capture a broad range of knowledge. Reasoning mechanisms are primarily retrieval-based, with emphasis on pattern recognition, data alignment, and summarization. Model generalization remains limited, operating primarily within predefined domains \cite{white2023future,pei2024leveraging,chen2023fine}. 

\textbf{Stage 2: Analogical Reasoning and Generalization.} This stage emphasizes the ability to draw connections and analogies across domains. Data requirements shift towards moderately diverse datasets that emphasize relationships and cross-domain patterns. Reasoning mechanisms incorporate relational reasoning and analogical thinking, enabling MLLMs to generalize effectively across domains. Applications include interdisciplinary problem-solving, transfer learning, and identifying cross-domain insight, reflecting a moderate increase in complexity and impact \cite{webb2023emergent,lewis2024evaluating}.

\textbf{Stage 3: Insightful Inference.} The third stage focuses on inferring deep insights from minimal and high-context data. Data requirements narrow to low-diversity, domain-specific datasets, allowing MLLMs to develop nuanced understanding. Reasoning mechanisms involve predictive reasoning and contextual interpretation, enabling the model to deduce complex outcomes. Generalization becomes highly context-specific, and applications include optimization and predictive modeling, making this stage highly impactful \cite{melko2024language,barman2025large}.

\textbf{Stage 4: Creative Hypothesis Generation.} In the final stage, MLLMs achieve the ability to generate innovative hypotheses and explore uncharted territories. Data requirements include highly diverse and synthetic datasets or simulation environments, fostering creativity. Reasoning mechanisms reach their highest complexity, involving generative reasoning and hypothesis exploration. Generalization becomes innovation-driven, synthesizing knowledge across fields. Applications at this stage have the highest impact, including proposing new theories, designing experiments, and driving scientific discovery \cite{xiong2024improving,qi2024large,pelletier2024explainable}.

\subsection{Data Heterogeneity Across Four Scientific Domains}
\label{sec:data_heterogeneity}
MLLMs are designed to process and integrate information from both textual and visual modalities, offering a versatile framework for handling complex scientific reasoning. However, the distinct nature of data across disciplines introduces unique challenges in model training and application. Appendix \ref{app:domain_data_differences} summarizes key differences in visual features for four scientific subjects within our scope.

Each subject presents unique challenges in data representation, with mathematical tasks primarily focusing on abstract symbols and formulas, while other subjects, particularly biology, require a mix of detailed real-world imagery and conceptual explanations. These differences necessitate domain-specific adaptations in how MLLMs process and understand multimodal data \cite{bai2024survey}.

\subsection{MLLM-based Scientific Reasoning}
\label{sec:mllm-based_reasoning}

\begin{figure}[t!]
    \centering
    \includegraphics[width=1 \linewidth]{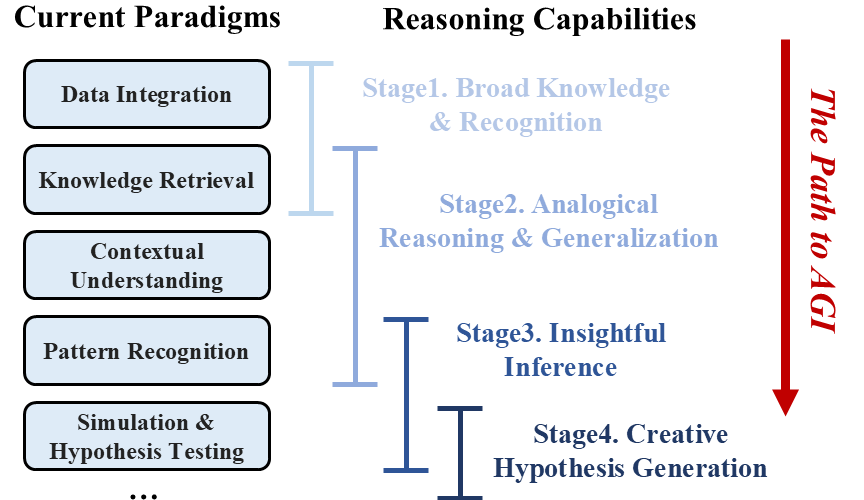}
    \caption{Overview of MLLM-based scientific reasoning paradigms and corresponding reasoning capabilities.}
    \label{fig:current_paradigms}
    \vspace{-6mm}
\end{figure}

As shown in Figure \ref{fig:current_paradigms}, current MLLM-based scientific reasoning can generally be divided into the following five paradigms, which progressively enhance the reasoning capabilities of MLLMs, ultimately moving towards AGI.

\textbf{Data Integration.} One of the primary strengths of MLLMs is their ability to integrate multimodal data from various scientific domains. For example, in physics, models can combine textual descriptions of a problem, such as Newton’s second law, with visual representations like force diagrams. The model can then reason about how different forces interact and predict motion \cite{barman2025large,sato2024exploring}. Similarly, in chemistry, MLLMs can combine chemical equations with 3D molecular structures, offering deeper insights into reaction mechanisms \cite{guo2023can}. This integrated approach allows MLLMs to handle intricate, often disjointed, data sources to generate coherent scientific explanations.

The ability to integrate and synthesize multimodal information enables the MLLM to solve complex problems more effectively. However, challenges arise when the visual data does not perfectly align with the textual explanation, potentially leading to misinterpretations \cite{zhang2025mathverse,lu2023mathvista,zhuang2024math}.

\textbf{Knowledge Retrieval.} A significant aspect of scientific reasoning is knowledge retrieval. In fields like physics and chemistry, the vast amount of scientific knowledge available - such as established theories, laws, or empirical data - can be overwhelming. Knowledge retrieval helps MLLMs access external knowledge bases, databases, and scientific literature to supplement their reasoning. For instance, when solving a chemistry problem, an MLLM could retrieve data from a chemical database to identify missing properties of substances or reactions that were not explicitly stated in the task \cite{prince2024opportunities,sze2024evaluation}. In addition, knowledge retrieval can aid MLLMs in bridging gaps between modalities. For example, an MLLM working on a biological problem may retrieve relevant studies from scientific papers to fill in missing knowledge, such as identifying unknown interactions between proteins or cells \cite{li2024benchmarking,li2025biomedrag}. 

This aspect of MLLMs’ reasoning ensures the model remains up-to-date with the latest discoveries and can apply a deeper layer of scientific knowledge in reasoning processes. However, challenges in accurately selecting and integrating relevant knowledge remain, particularly when sources of conflict are present \cite{fan2024survey}.


\textbf{Contextual Understanding.} Contextual understanding in scientific reasoning involves understanding not only the literal data presented but also the broader context in which it is used. MLLMs are capable of this by combining visual data, such as molecular structures in chemistry, with textual descriptions of chemical properties. This allows them to reason about potential interactions between molecules in a way that goes beyond simple matching \cite{liu2025integrating,horawalavithana2023scitune}. \citet{wang2024t} leverage Chain-of-Thought as teaching signals to train small models to perform reasoning in complicated scenarios.

This contextual capability is crucial in fields like biology, where visual images of biological processes must be linked with underlying theories to make accurate predictions. However, this capability can be limited when the model fails to integrate textual and visual information effectively, leading to errors in reasoning \cite{li2024multimodal}.

\textbf{Pattern Recognition.} In scientific reasoning, pattern recognition is a crucial skill that MLLMs excel at. MLLMs can detect patterns across different modalities, whether they are geometric in mathematics or experimental in chemistry. For instance, in biology, MLLMs can recognize cellular structures in images and relate them to known biological processes described in text, such as identifying mitochondria and correlating their function with energy production \cite{Luu2023BioinspiredLLMCL,kraus2024masked}. Additionally, an example of pattern recognition in mathematics could involve an MLLM matching visual representations of geometric figures with algebraic equations to find solutions to geometry problems \cite{mouselinos2024beyond}. This capability enhances model's ability to understand complex systems across disciplines, including identifying patterns in large datasets that might be intricate for manual analysis.

This skill allows MLLMs to be highly effective in analyzing multistep reasoning problems, which are often required in scientific disciplines \cite{qiao2024we,yan2024errorradar}. However, pattern recognition can be hindered by noisy or low-quality visual data, particularly in domains like biology, where image clarity is critical for correct interpretation \cite{zhang2024multimodal,ren2024pixellm}.

\textbf{Simulation and Hypothesis Testing.} MLLMs also possess the ability to perform simulation and hypothesis testing, a fundamental part of scientific reasoning \cite{qi2023large}. For example, in physics, MLLMs can simulate the effect of various forces on an object, predict outcomes, and validate those predictions against real-world data or experiments \cite{melko2024language,gao2024large,yu2024climsim}. This capacity allows MLLMs to conduct scientific inquiries in a manner akin to human researchers, testing hypotheses and refining conclusions \cite{morera2024foundation}. 

Despite these strengths, hypothesis testing is constrained by the quality \& quantity of available data, which in some scientific domains remains insufficient or incomplete. This limits the generalizability and reliability of MLLMs in tasks requiring deep, multistep reasoning \cite{xiong2024improving}.

\vspace{-2mm}
\section{Challenges}
\label{sec:challenges}

\begin{figure}[t!]
    \centering
    \includegraphics[width=1 \linewidth]{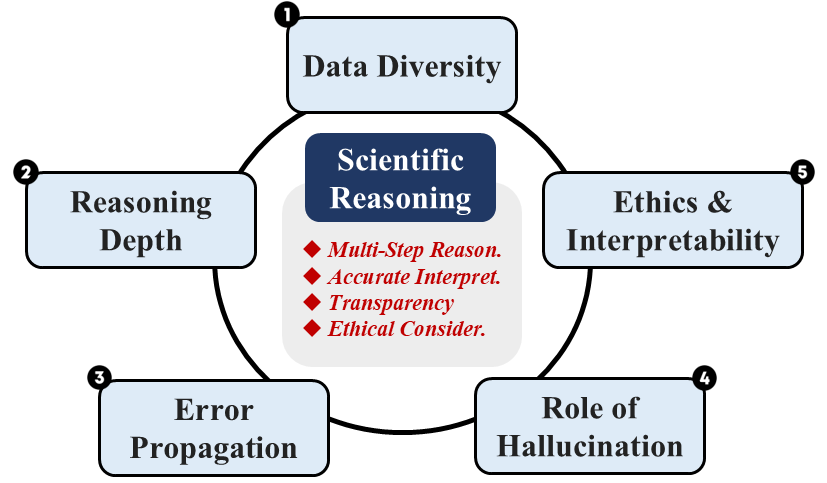}
    \caption{Challenges for MLLM-based scientific reasoning.}
    \label{fig:challenges}
    \vspace{-4mm}
\end{figure}

Even though MLLMs show substantial promise in solving scientific reasoning tasks, significant challenges remain. Based on the inherent characteristics of scientific reasoning - the need for multi-step inference and precise speculation, while ensuring transparency and ethicality - we further propose the following five key challenges (Figure \ref{fig:challenges}).


\textbf{Data Diversity.} Another challenge is the diversity of data across different scientific domains (summary of multi-domain scientific datasets can be seen in Appendix \ref{app:multidomain_scibench}). While mathematics is rich in textual data, such as equations and proofs, the availability of high-quality visual data is more limited \cite{qiao2024we,sun2024mm,liu2024mathllm,he2024olympiadbench}. In contrast, fields like chemistry and biology benefit from abundant visual data, such as molecular structures and microscopic images, but the corresponding textual descriptions may not always provide the depth required for comprehensive reasoning \cite{alampara2024probing,hocky2024connecting}. Without sufficient high-quality data for all modalities, model's ability to generalize across domains is compromised.

\textbf{Reasoning Depth.} MLLMs frequently struggle with tasks that require deep, multi-step reasoning, especially when abstract concepts are involved. In mathematics, for example, solving a theorem involves a series of logical steps that must be followed precisely \cite{chen2023theoremqa}. In physics, simulating complex systems, such as thermodynamics or quantum mechanics, requires a deep understanding of abstract principles and real-world conditions \cite{melko2024language}. MLLMs often fail to maintain this depth of reasoning, especially when applied to tasks involving intricate concepts or lengthy proof processes. This issue is particularly prevalent in fields where the complexity of reasoning extends beyond surface-level analysis and requires models to maintain rigorous logical consistency. Therefore, recent work has focused on two directions to improve the length of correct reasoning chains: the development of high-quality reasoning process datasets \cite{yan2024errorradar} and the introduction of process reward models \cite{skyworkopeno12024}.

\textbf{Error Propagation.} Error propagation is another significant challenge in multimodal reasoning. Errors in one modality, such as a misinterpreted graph or an unclear image, can propagate throughout the reasoning process, leading to incorrect conclusions \cite{li2024evaluating,yan2024errorradar,li2024ask}. For example, in a physics problem involving force vectors, an error in interpreting the vector diagram could lead to an incorrect calculation of the net force, which would then affect subsequent steps in the solution process \cite{jaiswal2024improving}. The risk of error propagation is particularly high when models are tasked with handling complex, multistep problems across multiple modalities. In particular, the impact of error propagation is especially acute in fields like physics and chemistry, where the accuracy of one step can influence the entire solution process. Small errors in initial data interpretation can lead to significant discrepancies in the final outcome \cite{xu2024ai,li2024bringing}.

\textbf{Role of Hallucinations.} One of the most complex challenges in leveraging MLLMs for scientific reasoning is determining whether hallucinations, the generation of information not grounded in the input data or knowledge base, are inherently harmful or potentially beneficial \cite{bai2024hallucination,liu2024survey}. While hallucinations are widely regarded as detrimental in factual tasks, their role in scientific reasoning is nuanced, particularly when considering the ultimate goal of advancing to Stage 4 of the research roadmap (\textit{i.e.}, Creative Hypothesis Generation) \cite{jiang2024survey}. In scientific reasoning, hallucinations can undermine trust and reliability by introducing inaccuracies in critical domains. For example, in physics, a hallucinated formula or principle might lead to invalid conclusions, while in chemistry, a fabricated reaction pathway could suggest impossible or even dangerous experiments. These inaccuracies not only hinder immediate problem-solving but also propagate errors if used as a basis for further research \cite{li2024deceptive,chakraborty2024hallucination,xu2024hallucination}. See details of hallucinations in scientific reasoning in Appendix \ref{app:hallucinations_for_reasoning}. 

\textbf{Ethical and Interpretability Issues.} Ethical concerns and model interpretability are major challenges when deploying MLLMs in high-stakes scientific domains, such as medical research or chemical engineering. MLLMs often lack transparency, making it difficult for users to understand how the model arrived at a particular conclusion \cite{alsaad2024multimodal}. Furthermore, ethical concerns arise when MLLMs are used to make decisions that could have significant consequences, such as in medical diagnoses or environmental impact assessments \cite{rahman2024survey}. In biology and medicine, the potential for biased reasoning in MLLMs, especially when trained on unbalanced datasets, could lead to harmful or misleading conclusions \cite{Stureborg2024LargeLM,wang2024fair}. An MLLM trained on biased medical data could fail to recognize critical symptoms in underrepresented populations, leading to erroneous diagnoses or treatment recommendations.

\vspace{-2mm}
\section{Discussion: What Next?}
\label{sec:discussion}

\begin{figure}[t!]
    \centering
    \vspace{-2mm}
    \includegraphics[width=1 \linewidth]{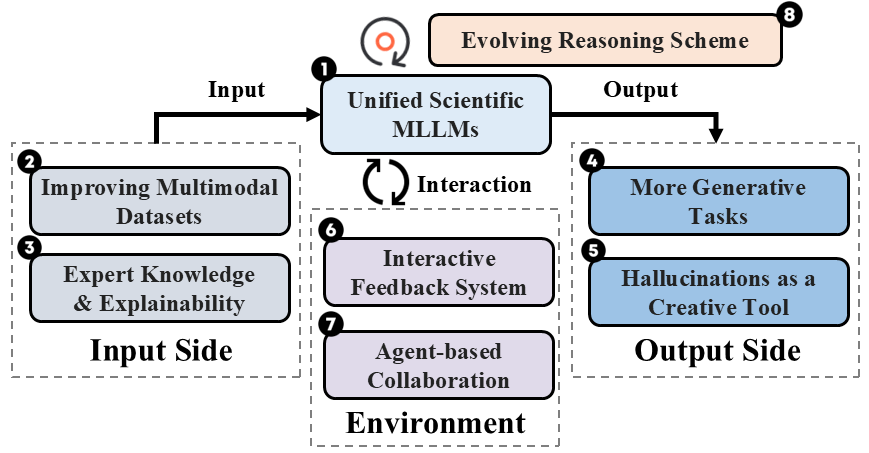}
    \caption{Eight prospects for the future of MLLMs in the field of multimodal scientific reasoning.}
    \label{fig:future_directions}
    \vspace{-4mm}
\end{figure}

Building on the challenges outlined in Section \ref{sec:challenges}, it is evident that while MLLMs hold great promise in advancing scientific reasoning, targeted solutions must be developed to address the limitations. Thus, we explore eight key perspectives for improving MLLMs in scientific reasoning (Figure \ref{fig:future_directions}).

\textbf{The Necessity of Unified Scientific MLLMs.}  
Although many domain-specific models have achieved remarkable performance in specialized scientific fields, exploring unified scientific MLLMs remains a critical pursuit \cite{taylor2022galactica}. Domain-specific models are optimized for particular areas (summary of scientific MLLMs can be seen in Appendix \ref{app:multimodal_scillm}), but they often lack the ability to integrate knowledge across disciplines \cite{wang2023survey,shi2024continual}. In contrast, a unified MLLM could facilitate interdisciplinary reasoning, leveraging connections between fields to tackle complex problems that require holistic understanding, \textit{e.g.,} climate change modeling \cite{nguyen2023climax} or biomedical research \cite{wang2023pre}.  

For example, a unified scientific MLLM could simultaneously analyze chemical reaction pathways and their biological implications, enabling breakthroughs in drug discovery \cite{oniani2024emerging,guan2024drug}. Similarly, it could integrate physics-based simulations with mathematical optimization to design more efficient renewable energy systems \cite{gao2024physically}. 

\textbf{Improving Multimodal Datasets.} A critical step in advancing MLLMs' capabilities is the improvement of multimodal datasets \cite{gadre2024datacomp,rahate2022multimodal,bayoudh2022survey}. Current datasets often lack the richness and variety required to train models effectively across disciplines. For instance, in chemistry, existing datasets may focus heavily on molecular structures without providing sufficient textual descriptions of reaction mechanisms \cite{cao2023instructmol}. Similarly, in biology, while there are abundant images of anatomical structures, these are often not paired with detailed descriptions of biological processes \cite{tang2023explainable,zhang2024benchmark}. By creating multimodal datasets that include both high-quality images and comprehensive textual descriptions across all domains, the model can better learn to correlate visual features with corresponding scientific explanations \cite{albalak2024survey,muennighoff2023scaling}. See discussion of integration with more modalities in Appendix \ref{app:more_modality}.

For example, in physics, datasets that combine experimental setups with corresponding theoretical explanations could help improve a model’s understanding of underlying principles, such as energy conservation or force interactions. This integration would facilitate a more robust training process, ensuring that models can handle a wider range of scientific reasoning tasks. As a concrete suggestion, developing domain-specific multimodal datasets that cover not only the subject but different teaching contexts (\textit{e.g.,} beginner vs. advanced materials) would help MLLMs generalize across varying complexity levels \cite{shi2023large}.

\textbf{Integrating Expert Knowledge and Explainability.} Integrating expert knowledge into MLLMs can significantly enhance their ability to reason accurately and logically \cite{pan2024unifying}. Expert knowledge, such as specialized theories in physics or established principles in biology, provides a framework within which the MLLM can operate. Additionally, by integrating causal reasoning into MLLMs, the models can better explain the relationships between variables, such as causes and effects, rather than simply identifying correlations \cite{xiong2024improvingcausal,jin2024cladder}. 

For example, in chemistry, integrating knowledge about chemical bonding, molecular dynamics, and reaction mechanisms can help guide the model’s predictions and interpretations \cite{zaki2024mascqa}. Additionally, enhancing the explainability of MLLMs is crucial for transparency \cite{dang2024explainable}. Ensuring that the model can justify its reasoning in a way that humans can easily understand, especially when making scientific claims, will increase its credibility and trustworthiness.

\textbf{Expanding Scientific Reasoning to Generative Tasks.}
While significant progress has been made in using MLLMs for problem-solving, error detection, and theorem proving \cite{yan2024survey}, the potential of these models in generative tasks remains underexplored. Generative tasks such as creating curriculum-aligned questions \cite{mulla2023automatic}, designing comprehensive syllabi \cite{hu2024teaching}, or enabling digital teaching assistants \cite{onu2024potential} are highly relevant to real-world applications, \textit{esp.} in educational contexts. 

For example, MLLMs could be used to generate topic-specific exam questions that align with diverse educational standards or assist in creating adaptive learning materials tailored to different proficiency levels \cite{denny2024desirable}. 

\textbf{Hallucinations as a Creative Tool.} Hallucinations may play a constructive role in fostering creativity and innovation, particularly in exploratory scientific tasks \cite{huang2023survey,li2024dawn}. At the frontier of Stage 4 (Section \ref{sec:roadmap}), MLLMs can hypothesize beyond existing knowledge, where generating plausible but unverified information might stimulate novel ideas. 

A critical challenge lies in striking the balance between mitigating harmful hallucinations and leveraging beneficial ones. This requires a fine-grained approach to model design, where hallucination mechanisms are carefully controlled based on task context. For foundational reasoning tasks, strict adherence to validated knowledge is essential, while exploratory tasks allow controlled deviations to inspire innovation \cite{jiang2024survey}.

\textbf{Interactive Feedback Systems.} Another powerful strategy involves the development of interactive feedback systems, which would allow MLLMs to engage with users dynamically, iterating on their answers based on user input or feedback \cite{abramson2022improving,shtarbanov2023sleeveio}. This interactive feature would not only enable models to adjust their reasoning during the problem-solving process but also allow them to ask clarifying questions, improving the overall user experience and enhancing the model’s output.

For example, in biology, a researcher could input a biological query related to an ecological model and receive initial results from the model. This back-and-forth interaction would provide a mechanism for error correction and refinement, ensuring that outputs align more closely with expert understanding \cite{pan2023automatically}. 

\textbf{Agent-based Collaboration.} A promising avenue for future development is agent-based collaboration, which involves the integration of multiple specialized agents working together to solve complex scientific problems \cite{guo2024large,wang2024survey}. Each agent could be tailored to handle specific scientific tasks, such as mathematical reasoning, chemical reaction prediction, or biological system analysis. These agents could communicate and collaborate with each other to cross-check information, validate hypotheses, and combine knowledge from their respective domains \cite{chen2023agentverse,yan2025mathagent}.

For instance, in a physics problem involving both mechanics and electromagnetism, an agent focused on classical mechanics could collaborate with an agent specialized in electromagnetism to deliver a comprehensive solution that accounts for the interactions between mechanical forces and electromagnetic fields. By building a system where multiple agents, each bringing a unique expertise, work collaboratively, MLLMs could approach complex scientific reasoning tasks with higher accuracy and robustness \cite{guo2024large,zhao2024expel}.

\textbf{Evolving Reasoning Schemes.} Current reasoning architectures in MLLMs remain constrained by opaque, monolithic designs that limit adaptability to diverse scientific domains \cite{besta2025reasoning}. Existing paradigms of Reasoning Large Models (RLMs) bifurcate into implicit RLMs (\textit{e.g.,} QwQ \cite{teamqwq}), where reasoning is embedded in model weights as a black box, and explicit RLMs (\textit{e.g.,} LLaMA-Berry \cite{zhang2024llama} \& o1 \cite{jaech2024o1}), which deploy structured reasoning strategies like Monte Carlo Tree Search (MCTS) or Beam Search with modular components. While explicit methods enable stepwise evaluation, their reliance on fixed templates and proprietary training schemes hinders reproducibility and domain-specific customization.

To advance scientific reasoning, future MLLMs should integrate three innovations: (i) dynamic reasoning structures (\textit{e.g.,} nested graphs) that adapt to multimodal inputs; (ii) process-based supervision with stepwise uncertainty metrics (\textit{e.g.,} token-level entropy) to refine domain-specific reasoning paths; and (iii) open-source, composable toolkits for hybrid training (\textit{i.e.,} Supervised Fine-Tuning + Reinforcement Learning phases) that decouple policy/value models, enabling collaborative, cost-efficient optimization across scientific disciplines \cite{besta2025reasoning}.
\section{Alternative Views}
\label{sec:alternative_view}
\vspace{-2mm}
We also discuss two key opposing perspectives and provide counterarguments to address these concerns in Appendix \ref{app:alternative_details} due to the space limit.

\section{Conclusion}
\label{conlcusion}
This position paper aims to emphasize the transformative potential of MLLMs in advancing scientific reasoning across diverse domains, including mathematics, physics, chemistry, and biology. Our key position is that MLLMs represent a significant step forward in enabling more comprehensive and accurate reasoning about scientific phenomena, bridging gaps between different types of data and reasoning methods. To support this stance, we reviewed the current state of MLLM applications in scientific reasoning, outlined key challenges, and proposed actionable insights.


\clearpage
\section*{Limitations}

While this position paper presents a comprehensive vision for the role of MLLMs in scientific reasoning, we acknowledge several limitations that define the boundaries of our current analysis and highlight avenues for future exploration.

\begin{itemize}
    \item \textbf{Focused Disciplinary Scope.} Our analysis is anchored in four core scientific disciplines: mathematics, physics, chemistry, and biology. While these fields are foundational and highly representative of multimodal reasoning challenges, they do not encompass the full spectrum of scientific inquiry. Future work could extend our proposed framework to other domains, such as earth sciences, materials science, and the social sciences, which present their own unique data modalities and reasoning paradigms.

    \item \textbf{Conceptual Nature of the Roadmap.} The proposed four-stage research roadmap is intended as a high-level conceptual framework to guide future development. We recognize that the progression between stages may not be strictly linear and the boundaries can be fluid. Establishing fine-grained, quantitative metrics to precisely benchmark an MLLM's position within this roadmap is a complex challenge that we leave for future research.
    
    \item \textbf{Focus on Model Capabilities over Human-AI Interaction.} Our discussion centers predominantly on the intrinsic reasoning capabilities of MLLMs. While we briefly touch upon interactive systems, a deeper exploration of the socio-technical dynamics—how these advanced models will effectively and ethically integrate into the daily workflows of human scientists and foster optimal human-AI collaboration—is beyond our current scope. This represents a rich and vital avenue for future work at the intersection of AI, HCI, and the philosophy of science.
\end{itemize}

\section*{Acknowledgments}
This work was supported by the Squirrel AI research fund; National Natural Science Foundation of China (Grant No.62506318); Guangdong Provincial Department of Education Project (Grant No.2024KQNCX028); Scientific Research Projects for the Higher-educational Institutions (Grant No.2024312096), Education Bureau of Guangzhou Municipality; Guangzhou-HKUST(GZ) Joint Funding Program (Grant No.2025A03J3957), Education Bureau of Guangzhou Municipality.

\bibliography{mllm4sci_position}


\clearpage
\newpage
\appendix
\hypersetup{linkcolor=black}
\etocdepthtag.toc{mtappendix}
\etocsettagdepth{mtchapter}{none}
\etocsettagdepth{mtappendix}{section}
\etocsettagdepth{mtappendix}{subsubsection}
\tableofcontents
\clearpage

{\large\textbf{Technical Appendices and Supplements}}

In this appendix, we first provide a clarification of our position's scope (Appendix \ref{app:scope_clarification}), detailing its novelty, the unique challenges MLLMs face in scientific reasoning compared to general tasks, the significance of cross-domain reasoning, and the integration of experimentation with our proposed roadmap. Then, in Appendix \ref{app:task_formulation}, we offer a formal formulation of the scientific reasoning task, while Appendix \ref{app:roadmap} elaborates on our four-stage research roadmap. Appendix \ref{app:domain_data_differences} discusses data heterogeneity across key scientific domains, and Appendix \ref{app:multidomain_scibench} reviews relevant multi-domain scientific reasoning benchmarks, with Appendix \ref{app:multimodal_scillm} providing an overview of current scientific (M)LLMs. In Appendix \ref{app:hallucinations_for_reasoning}, we present an in-depth analysis of hallucinations in scientific reasoning, including their types, existing mitigation strategies, and why they often fall short in scientific contexts. Appendix \ref{app:more_modality} further explores the necessity and technical approaches for integrating richer modalities such as audio, video, and 3D data. Finally, Appendix \ref{app:alternative_details} addresses alternative views and counterarguments, such as the preference for hyper-specialized models and concerns regarding MLLM reliability and explainability.

\section{Clarification of the Position Scope}
\label{app:scope_clarification}

\subsection{Novelty of Our Position}
\label{app:position_novelty}

This position paper presents a novel and compelling argument for the transformative potential of MLLMs in advancing the full spectrum of scientific reasoning, coupled with a unique, MLLM-centric four-stage roadmap towards this goal. While individual concepts such as applying AI to science or outlining stages of AI development have been discussed elsewhere, our work distinguishes itself through a specific synthesis, focus, and scope that addresses critical gaps in the current discourse:

\begin{enumerate}[leftmargin=*]
    \item \textbf{Primacy of Multimodality for Comprehensive Scientific Reasoning:} Our central thesis is that the inherent ability of MLLMs to integrate and reason over diverse data modalities is paramount for a holistic advancement in scientific reasoning. This contrasts sharply with existing literature that often focuses on unimodal approaches or narrower applications.
    \begin{itemize}
        \item For instance, comprehensive surveys on ``Scientific LLMs'' \cite{zhang2024comprehensive} predominantly analyze text-based models and their applications in scientific discovery, largely overlooking the rich multimodal nature of scientific data and reasoning. Our position explicitly champions the multimodal perspective as indispensable.
        \item Similarly, while works like \cite{reddy2024towards} discuss generative AI for ``scientific discovery,'' their scope is primarily confined to this (albeit important) sub-task. Our position encompasses a broader range of scientific reasoning capabilities crucial for overall scientific advancement, including foundational knowledge acquisition, commonsense scientific question answering, analogical reasoning across disciplines, and insightful inference from complex data – all areas where MLLMs offer unique advantages.
    \end{itemize}

    \item \textbf{A Dedicated Four-Stage Roadmap for MLLM-Driven Scientific Reasoning Towards AGI:} We propose a structured four-stage research roadmap (Broad Knowledge and Recognition, Analogical Reasoning and Generalization, Insightful Inference, and Creative Hypothesis Generation) that is specifically tailored to the evolving capabilities of MLLMs within \textit{all scientific reasoning scenarios}, ultimately aiming towards Artificial General Intelligence (AGI). This roadmap is distinct from others in its MLLM-centricity and breadth:
    \begin{itemize}
        \item It moves beyond general discussions on generative AI's potential in science \cite{morris2023scientists}, which, while insightful, did not offer a structured, MLLM-focused developmental pathway.
        \item It differs significantly from task-specific pipelines. For example, the highly valuable ``AI Scientist'' framework proposed by \cite{lu2024ai} outlines a three-stage pipeline (idea generation $\rightarrow$ experiment iteration $\rightarrow$ paper write-up) primarily designed for \textit{automated scientific discovery and specific generative outputs} like code and papers. Our roadmap is more encompassing: it situates such advanced discovery and generation capabilities (which we discuss in Section \ref{sec:discussion}, including applications like digital teaching assistants) within its sophisticated later stages (particularly Stage 4). Crucially, our roadmap also articulates the foundational and intermediate reasoning abilities (Stages 1-3) that are prerequisite for achieving such creative and autonomous scientific exploration with MLLMs. These earlier stages, focusing on multimodal data integration, knowledge retrieval, contextual understanding, and analogical generalization, are essential for building robust and reliable scientific reasoning systems.
    \end{itemize}
\end{enumerate}

In essence, the novelty of our position lies not in addressing these themes in isolation, but in the \textbf{holistic and integrated argument} that MLLMs, by virtue of their multimodal capabilities, are uniquely positioned to revolutionize scientific reasoning across its entire breadth. Furthermore, we provide a \textbf{dedicated, MLLM-specific, and progressive developmental trajectory} towards realizing this vision. This distinct stance, we believe, fills a crucial gap and warrants greater exposure and discussion within the machine learning community, aligning perfectly with the objectives of a Position Paper to highlight compelling arguments that can shape future research directions.

\subsection{Unique Challenges for MLLMs in Scientific Reasoning vs. General Challenges}
\label{app:challenges_comparisons}
While MLLMs face a spectrum of general challenges, their application to the domain of scientific reasoning introduces distinct and often amplified difficulties. We identify four key areas where the challenges for MLLMs in scientific reasoning diverge significantly from those encountered in more general applications:

\begin{enumerate}[leftmargin=*]
    \item \textbf{Demand for Rigor, Precision, and Verifiability:}
    In general tasks such as summarization or creative writing, MLLMs may employ approximate reasoning and provide plausible-sounding answers where occasional logical flaws or imprecision might be acceptable. However, scientific reasoning mandates an exceptionally high standard. It requires strict logical consistency, mathematical precision, and the ability to produce verifiable step-by-step derivations. The outputs must not only be correct but also demonstrably so, adhering to the rigorous validation principles of the scientific method.

    \item \textbf{Highly Specialized and Structured Multimodal Data:}
    General-purpose MLLMs are typically designed to handle diverse, often unstructured or semi-structured, multimodal data, with alignment efforts often focusing on broad semantic correspondence. In contrast, scientific reasoning necessitates the processing and deep understanding of highly structured, symbolic, and specialized data modalities. This includes, but is not limited to, complex chemical formulas, genetic sequences, intricate diagrams with specific notational conventions, and precise experimental data. Effective integration and reasoning over such specialized data require more than general semantic understanding; they demand the ability to parse and interpret domain-specific syntax and semantics.

    \item \textbf{The Critical Role of Causality and Mechanistic Understanding:}
    While MLLMs often excel at pattern recognition and identifying correlations within large datasets---valuable capabilities in many contexts---scientific reasoning demands a more profound level of understanding. A core objective of science is to move beyond mere correlation to infer causality and elucidate the underlying mechanisms that govern phenomena. MLLMs applied to science must therefore develop capabilities to not just describe \textit{what} happens, but to reason about \textit{why} and \textit{how} it happens, forming a crucial distinction from general pattern-matching tasks.

    \item \textbf{Nuanced Role and High Stakes of Hallucination:}
    Hallucinations, or the generation of factually incorrect or nonsensical information, are generally undesirable in any MLLM application as they lead to misinformation. However, in the context of scientific reasoning, factual hallucinations are particularly detrimental due to the high stakes involved; inaccuracies can misdirect research, lead to flawed conclusions, and erode trust. Paradoxically, there is also a unique potential role for \textit{controlled} generative capabilities---which might share some characteristics with hallucination if unconstrained---in the creative hypothesis generation phase. The challenge lies in fostering this creative, ``out-of-the-box'' thinking while strictly preventing uncontrolled factual inaccuracies, demanding a sophisticated balance not typically required in general applications.
\end{enumerate}

\subsection{Significance and Examples of Reasoning across Scientific Domains}
\label{app:significance_across_domains}

The capacity to synthesize knowledge and reason across disparate scientific domains is fundamental to addressing complex, multifaceted global challenges and accelerating the pace of discovery. Many scientific frontiers lie at the intersection of disciplines, requiring an integrated understanding of phenomena that transcend traditional boundaries. MLLMs are uniquely positioned to facilitate this interdisciplinary synergy by their inherent ability to process, correlate, and reason over diverse data types—textual, visual, numerical, and symbolic—from various fields. This capability can unlock novel insights and solutions that might remain obscured when viewed through a single disciplinary lens. Below, we illustrate this potential with representative examples:

\begin{enumerate}[label=\arabic*., leftmargin=*, itemsep=0.5em]
    \item \textbf{Drug Discovery and Precision Medicine:} MLLMs can integrate visual molecular structures, textual biomedical literature on biological pathways, and numerical patient genomic data. This allows for accelerated identification of promising drug candidates and prediction of patient-specific responses, paving the way for personalized medicine.

    \item \textbf{Materials Science and Engineering:} By analyzing material microstructure images (visual), crystal structure information (symbolic/visual), and textual data from scientific literature on synthesis and properties, MLLMs can rapidly predict characteristics of novel materials. This capability can guide experimental design and accelerate the discovery of materials with desired functionalities.

    \item \textbf{Climate Modeling and Earth System Science:} MLLMs can fuse satellite imagery (visual), numerical weather pattern data, and textual reports on atmospheric composition. This synthesis enables the construction of more comprehensive climate models, uncovering complex correlations and improving predictions of climate change impacts and extreme weather events.
\end{enumerate}

\subsection{Integration with Experimentation with Our Roadmap}
\label{app:position_novelty}
Experimentation is undeniably central to scientific advancement. While our four-stage roadmap (Broad Knowledge and Recognition, Analogical Reasoning, Insightful Inference, Creative Hypothesis Generation) focuses on the evolution of MLLM \textit{reasoning capabilities}, it inherently supports the experimental process, which we view as a sophisticated workflow leveraging these abilities.

For instance, designing novel experiments draws upon Creative Hypothesis Generation (Stage 4) and Insightful Inference (Stage 3), while interpreting complex experimental data relies heavily on Insightful Inference. Our framework explicitly addresses this in Section \ref{sec:mllm-based_reasoning} (``Simulation \& Hypothesis Testing''), where MLLMs contribute to generating testable hypotheses, designing \textit{in silico} experiments, and predicting outcomes---tasks demanding advanced inference and creative generation (Stages 3 \& 4). The progression through our stages equips MLLMs to assist in experimental design (Stages 1 \& 2), facilitate simulations (Stages 3 \& 4), enhance multimodal data analysis (Stage 3), and support iterative refinement of hypotheses and experiments (Stages 3 \& 4). Furthermore, our Discussion (Section \ref{sec:discussion}) on the future outlook of MLLMs (input, model, output, environment) naturally aligns with the components of scientific experimentation, where MLLMs process experimental inputs, perform reasoning, generate outputs like protocols or analyzed results, and operate within the scientific environment.

In essence, the capacity to support and augment scientific experimentation is an emergent outcome as MLLMs advance through the foundational reasoning capabilities outlined in our roadmap, making them integral to the experimental lifecycle.

\subsection{Quantitative Analysis of Multimodal Components}
\label{sec:appendix_quant}

To substantiate our position on the challenges and opportunities for MLLMs in scientific reasoning, this section provides a quantitative analysis of how MLLMs process and rely on different modalities, with a particular focus on the visual component. The analysis draws upon empirical experiment from two comprehensive benchmarks, \textbf{SciVerse} \cite{guo2025sciverse} and \textbf{MathVerse} \cite{zhang2025mathverse}, which systematically deconstruct problem formulations to isolate and evaluate the models' visual reasoning capabilities.

The central hypothesis investigated is that while MLLMs are designed to be multimodal, their performance in complex scientific and mathematical domains is disproportionately reliant on textual information, often failing to genuinely "see" or reason from diagrams. This creates a significant performance gap when crucial information is shifted from text to the visual modality, mirroring many real-world scenarios.

\begin{figure*}[t!]
    \centering
    \vspace{-2mm}
    \includegraphics[width=1 \linewidth]{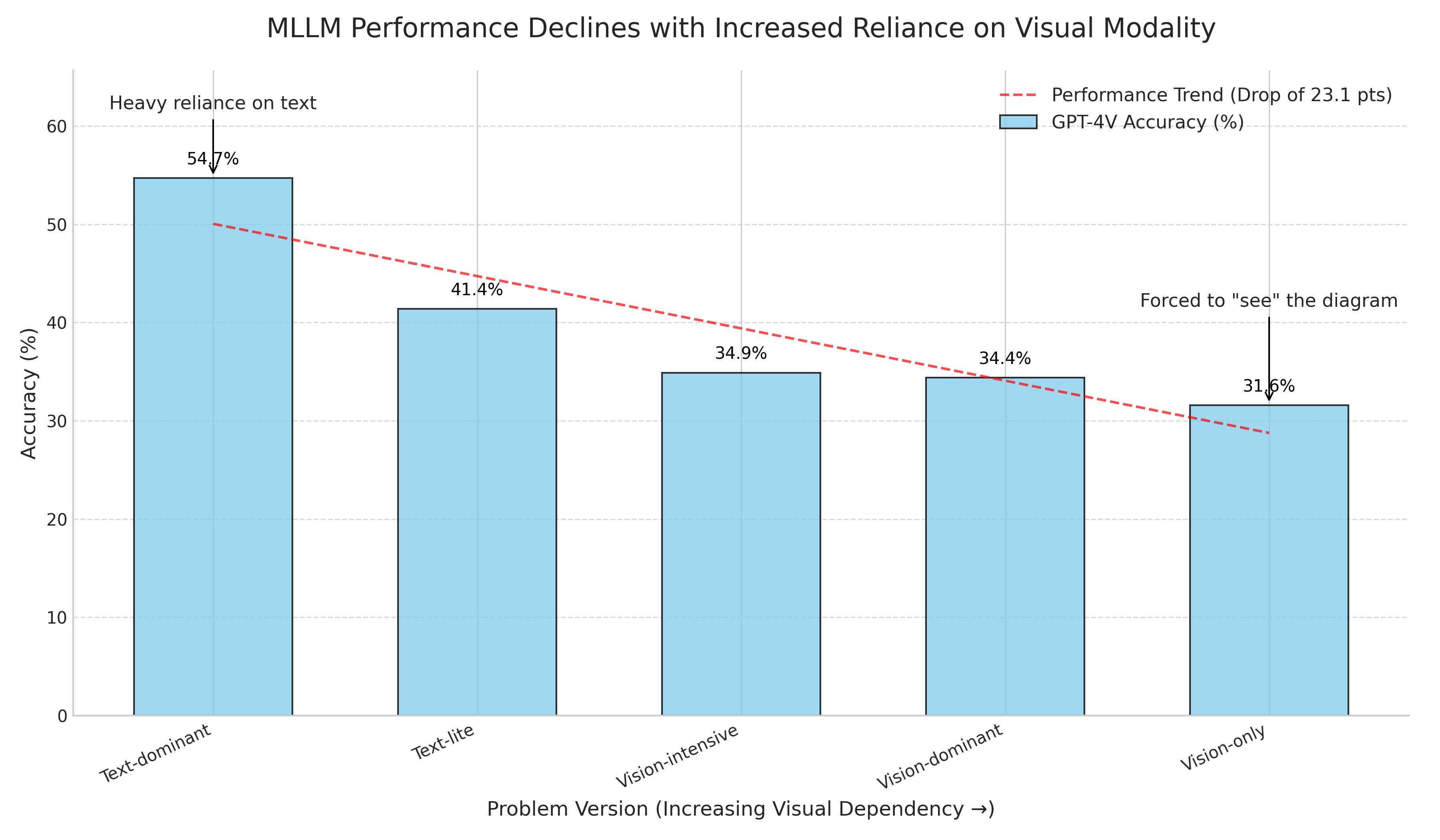}
    \caption{MLLM performance declines with increased reliance on visual modality from MathVerse benchmark.}
    \label{fig:mllm_visual_analysis}
\end{figure*}

\textbf{Analysis from SciVerse: General Scientific Reasoning.} The SciVerse benchmark evaluates MLLMs on scientific problems (Physics, Chemistry, Biology) by creating versions with varying degrees of visual dependency. The \texttt{Knowledge-free} version presents a standard problem with textual descriptions and a diagram. The \texttt{Vision-rich} version moves some conditional information from the text into the diagram, and the \texttt{Vision-only} version embeds the entire question within the visual input. As shown in Table~\ref{tab:sciverse_data}, the performance of even top-tier models like GPT-4o degrades as the reliance on visual perception increases.

\begin{table*}[t!]
  \centering
  \caption{Accuracy of MLLMs on SciVerse versions with increasing visual dependency. The performance drop from \texttt{Knowledge-free} to \texttt{Vision-only} highlights the challenge of visual information processing and reasoning.}
  \label{tab:sciverse_data}
  \begin{adjustbox}{max width=\textwidth}
  \begin{tabular}{lcccc}
    \toprule
    \textbf{Model} & \textbf{Knowledge-free (Acc \%)} & \textbf{Vision-rich (Acc \%)} & \textbf{Vision-only (Acc \%)} & \textbf{Perf. Drop (KF to VO)} \\
    \midrule
    GPT-4o               & 55.2 & 54.4 & 50.2 & \textbf{-5.0 pts} \\
    LLaVA-OneVision (7B) & 47.2 & 46.5 & 41.9 & \textbf{-5.3 pts} \\
    \bottomrule
  \end{tabular}
  \end{adjustbox}
  \vskip-0.5em
\end{table*}

The data reveals a consistent trend: when MLLMs are forced to extract information from diagrams rather than text, their accuracy declines. GPT-4o's accuracy drops by 5.0 percentage points, indicating that even state-of-the-art models struggle to interpret and integrate visually presented scientific conditions. This suggests that the visual encoder and cross-modal fusion mechanisms in current MLLMs are a significant bottleneck for general scientific reasoning.

\textbf{Analysis from MathVerse: Mathematical Reasoning.} The MathVerse benchmark provides an even more granular analysis for mathematical problems, a domain where diagrams are often structured and indispensable. It progressively shifts information from text to vision across six versions. For this analysis, we focus on the versions that best illustrate the visual dependency: \texttt{Text-dominant} (full text), \texttt{Text-lite} (redundant text removed), \texttt{Vision-intensive} (implicit properties in vision), \texttt{Vision-dominant} (key values in vision), and \texttt{Vision-only} (all in vision).

\begin{table*}[t!]
\centering
\caption{Accuracy of MLLMs across MathVerse versions, showing a progressive decline as problems become more vision-reliant.}
\label{tab:mathverse_data}
\begin{adjustbox}{max width=\textwidth}
\begin{tabular}{lcccccc}
    \toprule
    \textbf{Model} & \textbf{Text-dom (\%)} & \textbf{Text-lite (\%)} & \textbf{Vision-int (\%)} & \textbf{Vision-dom (\%)} & \textbf{Vision-only (\%)} & \textbf{Overall Drop} \\
    \midrule
    GPT-4V & 54.7 & 41.4 & 34.9 & 34.4 & 31.6 & \textbf{-23.1 pts} \\
    \midrule
    InternLM-XC2 (7B) & 22.3 & 17.0 & 15.7 & 16.4 & 11.0 & \textbf{-11.3 pts} \\
    \bottomrule
\end{tabular}
\end{adjustbox}
\vskip-0.5em
\end{table*}

The results, as detailed in Table~\ref{tab:mathverse_data} and Figure~\ref{fig:mllm_visual_analysis}, are striking. The steady and significant performance degradation as we move from \texttt{Text-dominant} to \texttt{Vision-only} provides compelling evidence that MLLMs heavily rely on textual "shortcuts." GPT-4V's accuracy plummets by over 23 percentage points, demonstrating a critical failure to parse essential conditions, properties, and questions from mathematical diagrams. MathVerse further reveals that some models, such as Qwen-VL-Max, astonishingly perform better without the diagram input at all (+5.1\%), treating the visual modality as a source of distraction rather than information.

The quantitative data from both SciVerse and MathVerse converges on a clear conclusion: a critical "vision-language gap" exists in the scientific reasoning capabilities of current MLLMs. Their strength lies predominantly in language processing, while their ability to perform deep, structured reasoning based on visual inputs—especially the information-rich diagrams common in science and math—remains underdeveloped. This quantitative analysis underscores the challenges highlighted in our position paper and reinforces the need for future research to focus on enhancing genuine multimodal understanding, moving beyond mere text-based reasoning with visual adornments.

\section{Formulation of Scientific Reasoning Task}
\label{app:task_formulation}
Mathematically, the scientific reasoning process can be modeled as an optimization task over a multimodal knowledge graph $G = (V, E)$, where $V$ denotes nodes (concepts, entities, or data points) and $E$ represents edges (relationships or interactions). Let $X_m$ represent the modality-specific input data, including mandatory modalities such as textual descriptions $X_t$ and visual representations $X_v$, as well as optional modalities like numerical data $X_n$ or other specialized inputs $X_o$. The goal is to predict or infer target outputs $Y$ based on a reasoning function $f_\theta$, parameterized by $\theta$, over $G$:
\[
Y = f_\theta(G, X_m) = f_\theta(V, E, X_t, X_v, X_n, X_o).
\]

\textbf{Mathematics:} Consider solving a geometry problem where $X_t$ provides the problem description, $X_v$ includes the geometric diagram, and $X_n$ optionally contains measurements or coordinates. The reasoning function $f_\theta$ integrates these inputs to infer the solution, such as identifying the area of a triangle.

\textbf{Physics:} In a physics experiment, $X_t$ describes the theoretical background, $X_v$ presents the experimental setup image, and $X_n$ includes sensor data such as velocity or temperature measurements. The function $f_\theta$ predicts the outcome or validates a hypothesis.

\textbf{Chemistry:} A multimodal analysis of a chemical reaction could include $X_t$ for the reaction mechanism, $X_v$ for molecular structure visualizations, and $X_o$ for spectroscopic data. The model predicts reaction yield or product properties.

\textbf{Biology:} When studying gene expression, $X_t$ describes the biological context, $X_v$ contains microscope images, and $X_n$ optionally includes numerical gene expression levels. The reasoning function predicts gene interactions or cellular behavior.

These examples illustrate how MLLMs can handle diverse inputs, integrating mandatory and optional modalities to perform complex scientific reasoning tasks.

\section{Four Phases of Research Roadmap}
\label{app:roadmap}

MLLMs have seen significant advancements in recent years, positioning their reasoning capabilities as a pivotal element on the pathway to achieving Artificial General Intelligence (AGI) \cite{wang2024exploring,yan2024survey,sun2023survey,jin2024reasoning,yan2024urbanclip,wei2024enhancing}. However, realizing AGI requires navigating a structured roadmap characterized by progressively complex reasoning tasks. This section outlines the Four Phases of Research Roadmap, each delineated by unique data requirements, reasoning mechanisms, generalization abilities, and impact. Table~\ref{tab:dimensions} provides a comparative overview of these phases.

\subsection{Phase 1: Broad Knowledge and Recognition}

In this phase, MLLMs prioritize high diversity and low specificity in data and knowledge. Tasks involve integrating vast and diverse datasets, emphasizing retrieval and alignment mechanisms. For example, MLLMs excel in synthesizing encyclopedic knowledge, aligning visual inputs (\textit{e.g.,} diagrams) with textual descriptions, and providing domain-specific insights. The generalization is limited to specific domains, resulting in low-impact applications like data retrieval and integration.

\subsection{Phase 2: Analogical Reasoning and Generalization}

This stage emphasizes medium diversity and specificity in data, leveraging contextual understanding to enable relational and analogical reasoning. For instance, MLLMs may draw analogies between chemical reaction pathways and electrical circuits, enhancing interdisciplinary insights. The models exhibit medium-level generalization across domains, allowing for moderate complexity tasks like explaining scientific phenomena using cross-domain analogies.

\subsection{Phase 3: Insightful Inference}

As reasoning tasks grow in complexity, MLLMs in this phase focus on low diversity but high specificity datasets. Predictive reasoning mechanisms are central, enabling context-specific inferences. For example, a model might predict the behavior of a physical system under certain constraints or optimize complex processes like material design. These capabilities lead to high-impact applications, including scientific optimization and inferential problem-solving.

\subsection{Phase 4: Creative Hypothesis Generation}

The final phase demands both high diversity and high creativity in data. Generative reasoning mechanisms empower MLLMs to propose innovative solutions and simulate hypotheses. For instance, models might design novel molecules for drug discovery or hypothesize ecological models for sustainable ecosystems. This phase represents very high-impact applications, bridging the gap between scientific discovery and innovation.

\subsection{Summary}

In summary, the Four Phases of Research Roadmap reflect the increasing complexity and potential of MLLMs in scientific reasoning. While current MLLMs have demonstrated impressive capabilities, they remain far from achieving AGI \cite{mumuni2025large,wang2024lighthouse,feng2024far,fei2022towards}. The community must continue to explore advanced reasoning abilities, fostering collaboration and innovation along this roadmap to address the challenges of AGI-driven scientific reasoning.
\begin{table*}[t]
  \centering
  \caption{Comparison across four stages along key dimensions.}
  \begin{adjustbox}{max width=\textwidth}
  \begin{tabular}{ccccc}
    \toprule
    \textbf{Dimension} &
    \makecell{\textbf{Broad Knowledge}\\\textbf{and Recognition}} &
    \makecell{\textbf{Analogical Reasoning}\\\textbf{and Generalization}} &
    \makecell{\textbf{Insightful}\\\textbf{Inference}} &
    \makecell{\textbf{Creative Hypothesis}\\\textbf{Generation}} \\
    \midrule
    \makecell{\textbf{Data and Knowledge}\\\textbf{Requirements}} &
    \makecell{High diversity,\\low specificity} &
    \makecell{Medium diversity,\\medium specificity} &
    \makecell{Low diversity,\\high specificity} &
    \makecell{High diversity,\\high creativity} \\
   \hline 
   \makecell{\textbf{Reasoning}\\\textbf{Mechanisms}} &
    \makecell{Low complexity\\(retrieval, alignment)} &
    \makecell{Medium complexity\\(relational, analogical)} &
    \makecell{High complexity\\(predictive)} &
    \makecell{Very high complexity\\(generative)} \\
    \hline 
    \makecell{\textbf{Model}\\\textbf{Generalization}} &
    \makecell{Low\\(domain-specific)} &
    \makecell{Medium\\(cross-domain)} &
    \makecell{High\\(context-specific)} &
    \makecell{Very high\\(innovative)} \\
    \hline 
    \makecell{\textbf{Applications}\\\textbf{and Impact}} &
    \makecell{Low impact\\(retrieval, integration)} &
    \makecell{Medium impact\\(interdisciplinary insights)} &
    \makecell{High impact\\(optimization, inference)} &
    \makecell{Very high impact\\(discovery, innovation)} \\
    \bottomrule
  \end{tabular}
  \end{adjustbox}
  \label{tab:dimensions}
\end{table*}

\section{Data Differences among Four Domains}
\label{app:domain_data_differences}

The reasoning capabilities of MLLMs make them particularly well-suited for processing heterogeneous multimodal data \cite{pattnayak2024survey,li2024benchsurvey,li2024benchsurvey2}. By integrating and analyzing diverse data formats such as text, images, and structured information, MLLMs enable more comprehensive insights across various scientific disciplines. This section highlights the unique data characteristics of four domains: mathematics, physics, chemistry, and biology, as summarized in Table~\ref{tab:subject_features}.

\subsection{Mathematics: Structured Abstraction}

Mathematical data is characterized by its symbolic equations, graphs, and geometric figures. These data types are highly structured and abstract, requiring precise interpretation and manipulation. Visual features such as coordinate axes and geometric shapes often complement formal textual elements like equations and proofs. The integration of these modalities enables MLLMs to solve complex mathematical problems and support theorem proving \cite{gairmathabel,yan2024survey,Azerbayev2023LlemmaAO,li2024evaluating}.

\subsection{Physics: Real-world Dynamics}

Physics data encompasses diagrams (\textit{e.g.,} vector and circuit diagrams), graphs, and descriptions of real-world phenomena. Its data structure reflects system dynamics and real-world applications, combining descriptive text with visual representations like force vectors or particle motion. MLLMs leverage these multimodal inputs to model physical systems and predict outcomes under varying conditions \cite{jaiswal2024improving,barman2025large,yu2024climsim}.

\subsection{Chemistry: Molecular and Symbolic}

Chemistry relies heavily on molecular structures, reaction pathways, and the periodic table. The data is both symbolic and structural, involving visual elements such as 3D molecular models and reaction schemes. Textural features include chemical equations and reaction mechanisms. MLLMs facilitate understanding chemical interactions and even predicting new compounds or reaction outcomes \cite{alamparamacbench,miret2024llms,mirza2024large,ramos2025review,m2024augmenting}.

\subsection{Biology: Conceptual Complexity}

Biological data is diverse, covering anatomical diagrams, cellular structures, and ecological models. Its structure is conceptual and often involves biological and anatomical representations. Visual inputs like cell diagrams and ecosystem models are paired with textual descriptions of processes and interactions. MLLMs support tasks such as identifying biological patterns and simulating ecological dynamics \cite{kraus2024masked,Luu2023BioinspiredLLMCL,huang2024adapting,wang2023pre,Liu2023GITMolAM,tang2023explainable}.

\subsection{Summary and Future Directions}

In summary, MLLMs demonstrate strong reasoning capabilities across mathematics, physics, chemistry, and biology by integrating diverse multimodal data. However, their potential is not confined to these four disciplines. Future research should explore reasoning capabilities in broader domains such as geospatial analysis \cite{roberts2024charting,mai2024opportunities,hao2024urbanvlp,zhong2024urbancross} and coding \cite{li2024mmcode,guo2024deepseekcode,di2024codefuse,hui2024qwen2code}, which demand advanced generalization abilities. Such endeavors are crucial for achieving the comprehensive reasoning capabilities envisioned in the AGI roadmap.
\begin{table*}[t]
  \centering
  \caption{Comparison of data features among four domains within the scope of this position paper.}
  \begin{adjustbox}{max width=\textwidth}
  \begin{tabular}{ccccc}
    \toprule
    \textbf{Feature/Domain} &
    \textbf{Mathematics} &
    \textbf{Physics} &
    \textbf{Chemistry} &
    \textbf{Biology} \\
    \midrule
    \textbf{Data Types} &
      \makecell{Symbolic equations,\\ graphs, geometric figures} &
      \makecell{Diagrams (vector, circuit),\\ graphs, real-world phenomena} &
      \makecell{Molecular structures,\\ reaction pathways,\\ periodic table} &
      \makecell{Anatomical diagrams,\\ cellular structures,\\ ecological models} \\
      \midrule
    \textbf{Data Structure} &
      \makecell{Abstract, highly structured,\\ formal notation} &
      \makecell{Real-world applications,\\ system dynamics} &
      \makecell{Symbolic and molecular,\\ structural} &
      \makecell{Conceptual, biological,\\ and anatomical} \\
      \midrule
    \textbf{Visual Features} &
      \makecell{Coordinate axes,\\ geometric shapes} &
      \makecell{Diagrams e.g., force vectors,\\ particle motion} &
      \makecell{3D molecular models,\\ reaction schemes} &
      \makecell{Photos, cell diagrams,\\ ecosystem models} \\
      \midrule
    \textbf{Textural Features} &
      \makecell{Equations, proofs,\\ formal definitions} &
      \makecell{Descriptive, experimental\\ setups, theories} &
      \makecell{Chemical equations,\\ reaction mechanisms} &
      \makecell{Biological processes,\\ ecological interactions} \\
    \bottomrule
  \end{tabular}
  \end{adjustbox}
  \label{tab:subject_features}
  \vskip-0.5em
\end{table*}

\section{Multi-Domain Scientific Reasoning Benchmark}
\label{app:multidomain_scibench}
The emergence of multi-domain scientific reasoning benchmarks has played a pivotal role in advancing AI models' ability to reason across diverse scientific domains. These benchmarks vary in their focus on multimodal integration, educational levels, and the comprehensiveness of domain coverage, as summarized in Table~\ref{tab:multi_domain_bench}.

\textbf{Comprehensive Domain Coverage:} Several benchmarks, such as \textbf{MMMU-Pro}, \textbf{CMMMU}, and \textbf{SciEval}, provide extensive domain coverage, including mathematics, physics, chemistry, and biology. These benchmarks target diverse educational contexts, ranging from primary education to PhD-level tasks, ensuring broad applicability. Notably, \textbf{MMMU-Pro} and \textbf{SciBench} have become instrumental for college-level evaluation, while \textbf{CMMU} extends its scope to younger learners, addressing a critical gap in foundational scientific reasoning.

\textbf{Multimodal Reasoning:} The integration of multimodal capabilities has become a defining feature of contemporary benchmarks. Over 60\% of the surveyed benchmarks, such as \textbf{EXAMS-V}, \textbf{SciBench}, and \textbf{ScienceQA}, incorporate multimodal tasks that involve textual, visual, and symbolic reasoning. These tasks reflect real-world problem-solving scenarios where multiple modalities interact, making them essential for evaluating the holistic reasoning capabilities of advanced AI models.

\textbf{Specialized vs. General Benchmarks:} While benchmarks like \textbf{OlympiadBench} and \textbf{OlympicArena} are tailored to high-stakes competitions, their narrow focus on specific tasks, such as Olympiad-level challenges, limits their generalizability. Conversely, resources like \textbf{AGIEval} and \textbf{SciEval} aim to provide a broader evaluation of scientific reasoning, covering multiple domains across various educational levels. However, their lack of multimodal integration highlights the need for more versatile benchmarks.


\textbf{Future Directions:} The future of multi-domain scientific reasoning benchmarks lies in developing unified frameworks that combine multimodal reasoning, domain comprehensiveness, and adaptability across educational levels. Incorporating elements like interactive feedback systems and collaborative reasoning tasks will bridge the gap between theoretical evaluation and practical applications, fostering the next generation of scientific reasoning models.
\begin{table*}[t]
\caption{Comparison of multi-domain scientific reasoning benchmarks.}
\label{tab:multi_domain_bench}
\centering
\begin{adjustbox}{max width=\textwidth}
\begin{tabular}{ccccccccc}
\toprule
\multirow{2}{*}{\textbf{Paper}} &
\multirow{2}{*}{\makecell{\textbf{Organization}}} &
\multirow{2}{*}{\makecell{\textbf{Venue}}} &
\multirow{2}{*}{\makecell{\textbf{Multimodal}}} &
\multirow{2}{*}{\makecell{\textbf{Education}\\\textbf{Level}}} &
\multicolumn{4}{c}{\textbf{Domain(s)}} \\
\cmidrule(lr){6-9}
 & & & & & \textbf{Math} & \textbf{Physics} & \textbf{Chemistry} & \textbf{Biology} \\
\midrule
\textbf{MMMU-Pro}~\cite{yue2024mmmupro} 
  & CMU 
  & ACL'25 
  & \checkmark 
  & College 
  & \checkmark 
  & \checkmark 
  & \checkmark 
  & \checkmark 
\\
\textbf{SciVerse}~\cite{guo2025sciverse} 
  & CUHK 
  & ACL Findings'25
  & \checkmark 
  & General 
  &  
  & \checkmark 
  & \checkmark 
  & \checkmark
\\
\textbf{MMSciBench}~\cite{ye2025mmscibench} 
  & Yale University
  & ACL Findings'25
  & \checkmark 
  & High 
  & \checkmark 
  & \checkmark 
  & 
  & 
\\
\textbf{EMMA}~\cite{hao2025can} 
  & UESTC 
  & ICML'25
  & \checkmark 
  & General 
  & \checkmark 
  & \checkmark 
  & \checkmark 
  & 
\\
\textbf{LLM-SRBench}~\cite{shojaee2025llm} 
  & Virginia Tech 
  & ICML'25
  &  
  & General
  &  
  & \checkmark 
  & \checkmark 
  & \checkmark
\\
\textbf{CURIE}~\cite{cui2025curie} 
  & Google 
  & ICLR'25
  & \checkmark 
  & General 
  &  
  & \checkmark 
  & \checkmark 
  & \checkmark
\\
\textbf{ScienceAgentBench}~\cite{chen2024scienceagentbench} 
  & OSU
  & ICLR'25 
  &  
  & General 
  & 
  & 
  & \checkmark 
  & \checkmark
\\
\textbf{Scientist-Bench}~\cite{tang2025airesearcher} 
  & HKU
  & NeurIPS'25 
  &  
  & General 
  & \checkmark 
  & \checkmark 
  & \checkmark 
  & \checkmark
\\
\textbf{MMVU}~\cite{zhao2025mmvumeasuring} 
  & Yale University 
  & CVPR'25
  & \checkmark 
  & College
  &  
  & \checkmark 
  & \checkmark 
  & \checkmark
\\
\textbf{SGI-Bench}~\cite{xu2025probing} 
  & Shanghai AI Lab
  & arXiv'25
  & \checkmark 
  & General
  & \checkmark
  & \checkmark 
  & \checkmark 
  & \checkmark
\\
\textbf{SDE}~\cite{song2025evaluating} 
  & Deep Principle
  & arXiv'25
  & 
  & General
  &  
  & \checkmark 
  & \checkmark 
  & \checkmark
\\
\textbf{SciBench}~\cite{wang2023scibench} 
  & UCLA 
  & ICML'24 
  & \checkmark 
  & College 
  & \checkmark 
  & \checkmark 
  & \checkmark 
  & 
\\
\textbf{MMMU}~\cite{yue2024mmmu} 
  & \makecell{University of Waterloo \\ \& Ohio State University} 
  & CVPR'24 
  & \checkmark 
  & College 
  & \checkmark 
  & \checkmark 
  & \checkmark 
  & \checkmark 
\\
\textbf{EXAMS-V}~\cite{das2024exams} 
  & MBZUAI 
  & ACL'24
  & \checkmark 
  & General 
  & \checkmark 
  & \checkmark 
  & \checkmark 
  & \checkmark 
\\
\textbf{ArxivQA/Cap}~\cite{li2024multimodalarxiv} 
  & HKU
  & ACL'24
  & \checkmark 
  & General 
  & \checkmark 
  & \checkmark 
  & \checkmark 
  & \checkmark 
\\
\textbf{OlympiadBench}~\cite{he2024olympiadbench} 
  & Tsinghua University
  & ACL'24 
  & \checkmark
  & Competition 
  & 
  & 
  & \checkmark 
  & \checkmark 
\\
\textbf{SceMQA}~\cite{liang2024scemqa} 
  & University of Notre Dame
  & ACL Short'24 
  & \checkmark
  & College
  & \checkmark 
  & \checkmark 
  & \checkmark 
  & \checkmark 
\\
\textbf{OlympicArena}~\cite{huang2024olympicarena} 
  & Shanghai Jiaotong University
  & NeurIPS'24 
  & 
  & Competition 
  & \checkmark 
  & \checkmark 
  & \checkmark 
  & \checkmark 
\\
\textbf{SciEval}~\cite{sun2024scieval} 
  & Shanghai Jiaotong University 
  & AAAI'24 
  & 
  & General 
  & \checkmark 
  & \checkmark 
  & \checkmark 
  & \checkmark 
\\
\textbf{CMMU}~\cite{he2024cmmu} 
  & \makecell{Beijing Academy of \\ Artificial Intelligence}
  & IJCAI'24 
  & \checkmark 
  & \makecell{Primary/\\Middle/High} 
  & \checkmark 
  & \checkmark 
  & \checkmark 
  & \checkmark 
\\
\textbf{AGIEval}~\cite{zhong-etal-2024-agieval} 
  & Microsoft 
  & NAACL'24 
  & 
  & \makecell{High/College} 
  & \checkmark 
  & \checkmark 
  & \checkmark 
  & \checkmark 
\\
\textbf{CMMLU}~\cite{li2024cmmlu} 
  & MBZUAI 
  & ACL Findings'24
  & 
  & High/College 
  & \checkmark 
  & \checkmark 
  & \checkmark 
  & \checkmark 
\\
\textbf{MMWorld}~\cite{he2024mmworld} 
  & UCSC 
  & arXiv'24 
  & \checkmark 
  & General
  & \checkmark 
  & \checkmark 
  & \checkmark 
  & \checkmark 
\\
\textbf{CMMMU}~\cite{zhang2024cmmmu} 
  & HKUST 
  & arXiv'24 
  & \checkmark 
  & College 
  & \checkmark 
  & \checkmark 
  & \checkmark 
  & \checkmark 
\\
\textbf{M4U}~\cite{wang2024m4u} 
  & Chinese Academy of Sciences
  & arXiv'24 
  & \checkmark 
  & College 
  & \checkmark 
  & \checkmark 
  & \checkmark 
  & \checkmark 
\\
\textbf{MMSci}~\cite{li2024mmsci} 
  & UCSB
  & arXiv'24 
  & \checkmark 
  & PhD 
  & 
  & \checkmark 
  & \checkmark 
  & \checkmark
\\
\textbf{LitQA2}~\cite{skarlinski2024language} 
  & FutureHouse Inc.
  & arXiv'24 
  &  
  & General 
  & \checkmark
  & \checkmark
  & \checkmark 
  & 
\\
\textbf{VisScience}~\cite{jiang2024visscience} 
  & Tsinghua
  & arXiv'24 
  & \checkmark  
  & \makecell{Primary/\\Middle/High}  
  & 
  & 
  & \checkmark 
  & \checkmark
\\
\textbf{JEEBench}~\cite{arora2023have} 
  & \makecell{Microsoft \\ \& UC Berkeley} 
  & EMNLP'23 
  & 
  & General 
  & \checkmark 
  & \checkmark 
  & \checkmark 
  &  
\\
\textbf{M3Exam}~\cite{zhang2023m3exam} 
  & Alibaba
  & NeurIPS'23 
  & \checkmark 
  & \makecell{Primary/\\Middle/High} 
  & \checkmark
  & 
  & \checkmark 
  &  
\\
\textbf{LitQA}~\cite{lala2023paperqa} 
  & FutureHouse Inc.
  & arXiv'23 
  &  
  & General 
  & 
  & 
  & \checkmark 
  & \checkmark
\\
\textbf{ScienceQA}~\cite{lu2022learn} 
  & UCLA 
  & NeurIPS'22 
  & \checkmark 
  & \makecell{Primary/High} 
  &  
  & \checkmark 
  & \checkmark 
  & \checkmark 
\\
\textbf{IconQA}~\cite{lu2021iconqa} 
  & UCLA 
  & NeurIPS'21 
  & \checkmark 
  & General 
  & \checkmark 
  & \checkmark 
  &  
  &  
\\
\textbf{MMLU}~\cite{hendrycks2021mmlu} 
  & UC Berkeley 
  & ICLR'21 
  & 
  & General 
  & \checkmark 
  & \checkmark 
  & \checkmark
  & \checkmark
\\
\bottomrule
\end{tabular}
\end{adjustbox}
\vskip-0.5em
\end{table*}

\section{Details of Hallucinations in Scientific Reasoning}
\label{app:hallucinations_for_reasoning}
Since a position paper must balance scope breadth and technical depth, we did not extensively elaborate on each specific challenge in the main text. However, we provide a more detailed discussion on hallucinations here, as they present a nuanced challenge particularly critical to the advancement of MLLMs in scientific reasoning. While often detrimental, understanding and managing hallucinations is key, especially considering the aspirational "Creative Hypothesis Generation" stage of our roadmap.

\begin{figure}[t!]
    \centering
    \vspace{-2mm}
    \includegraphics[width=1 \linewidth]{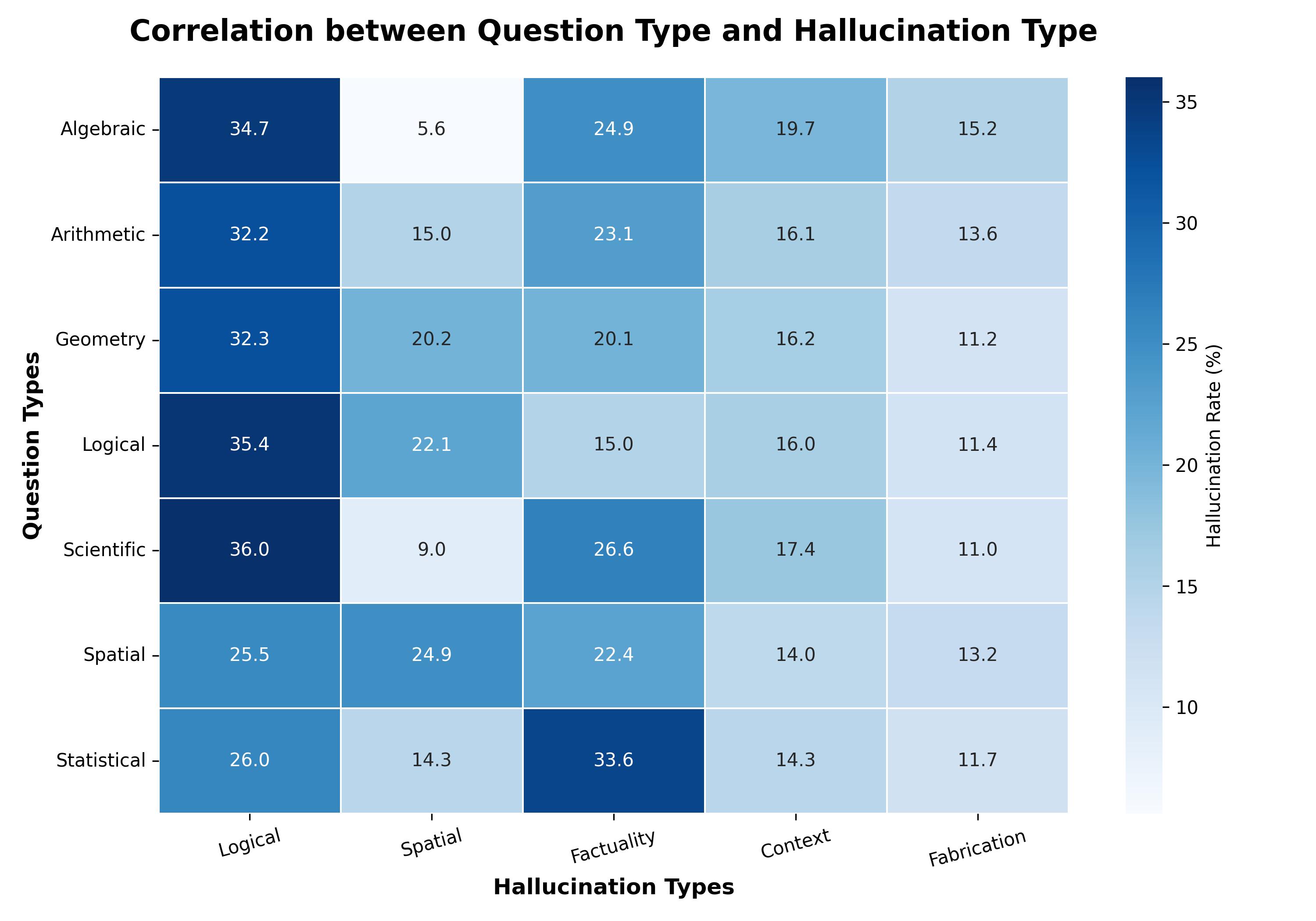}
    \caption{Eight prospects for the future of MLLMs in the field of multimodal scientific reasoning.}
    \label{fig:visualized_heatmap_hallucination}
    \vspace{-4mm} 
\end{figure}

\begin{figure}[t!]
    \centering
    \vspace{-2mm}
    \includegraphics[width=1 \linewidth]{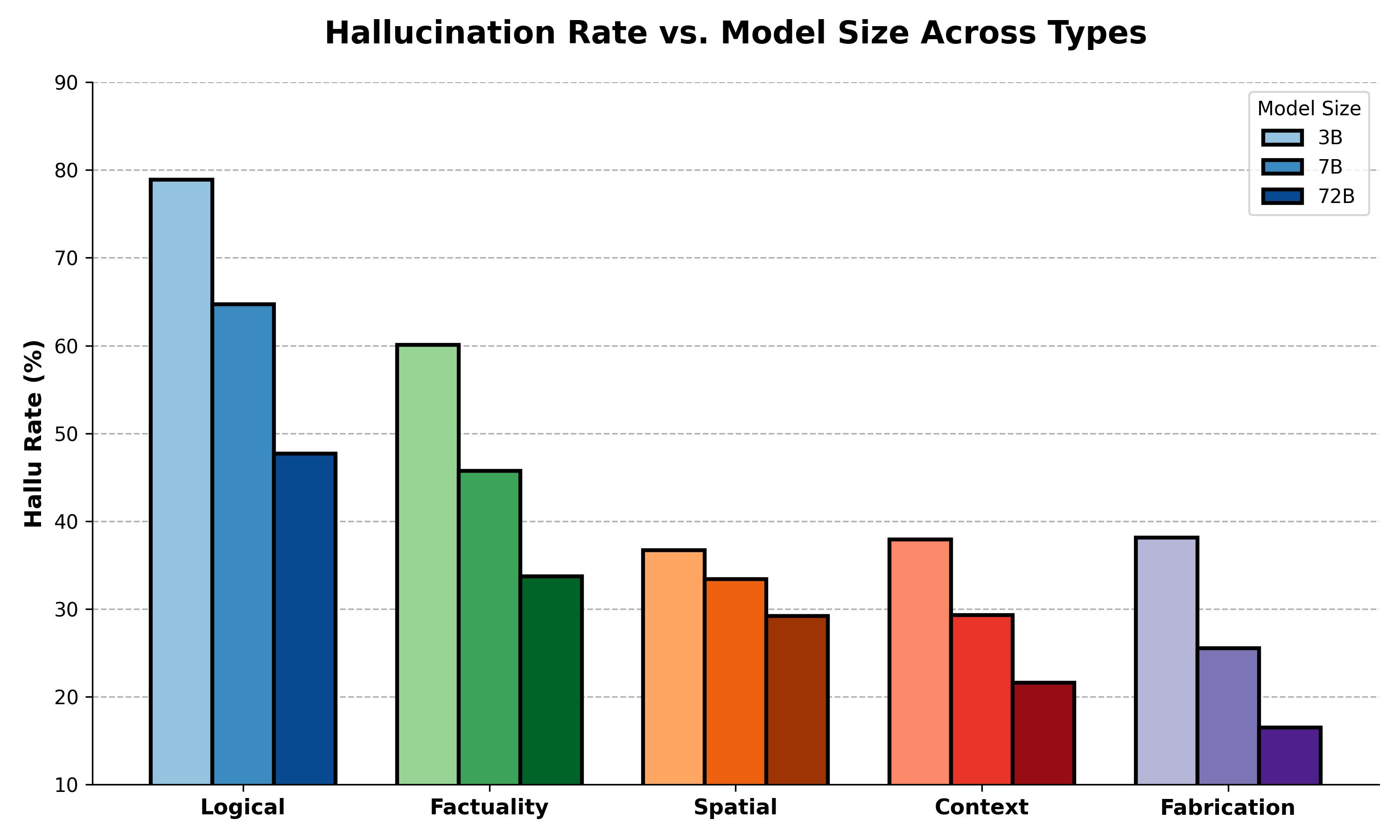}
    \caption{Eight prospects for the future of MLLMs in the field of multimodal scientific reasoning.}
    \label{fig:visualized_bar_hallucination}
    \vspace{-4mm}
\end{figure}

Recent studies on multimodal reasoning hallucination, such as MIRAGE \cite{dong2025mirage}, reveal complex and nuanced patterns that challenge one-size-fits-all mitigation strategies. As illustrated in Figure \ref{fig:visualized_heatmap_hallucination}, the distribution of hallucination types is strongly correlated with the nature of the scientific task; for instance, logical hallucinations are pervasive across most question types, whereas factuality hallucinations are more prominent in knowledge-intensive domains like statistics, and spatial hallucinations are uniquely problematic in geometry and spatial tasks. Furthermore, Figure  \ref{fig:visualized_bar_hallucination} demonstrates that while increasing model size can effectively reduce logical and fabrication hallucinations, it provides only marginal improvements for spatial hallucinations, indicating that this type of error stems from a deeper visual reasoning deficit that cannot be solved by simple scaling. These findings underscore that the hallucination categories defined for multimodal scientific reasoning, which isolate reasoning failures from perceptual errors, are distinct and behave differently from general-purpose hallucination concepts. Their unique characteristics and resistance to conventional scaling solutions cannot be fully captured by broader definitions, thus necessitating the more targeted, fine-grained hallucination classification.

\subsection{Different Types of Hallucinations in Scientific Reasoning} 

Hallucinations in MLLMs, broadly defined as the generation of information not grounded in the input data or established knowledge, manifest in several ways pertinent to scientific reasoning:

\begin{itemize}[leftmargin=*]
    \item \textbf{Factual Hallucination:} This is the most straightforward type, where the MLLM generates incorrect or non-existent facts, such as an incorrect physical constant, a misremembered chemical property, a wrong date for a discovery, or a fabricated citation. In science, where precision is paramount, factual hallucinations can lead to fundamentally flawed conclusions.
    \item \textbf{Conceptual Hallucination:} This involves the misrepresentation or misapplication of scientific concepts, theories, or principles. An MLLM might, for example, confuse correlation with causation in a biological context, misapply Newton's laws in a scenario where relativistic effects are dominant, or incorrectly define a mathematical theorem's conditions.
    \item \textbf{Relational Hallucination:} This occurs when an MLLM fabricates or misrepresents relationships between entities, variables, or concepts. It might invent a non-existent interaction between two proteins, incorrectly link a gene to a disease without evidence, or propose a flawed causal chain in a physical process.
    \item \textbf{Explanatory Hallucination:} Here, the MLLM generates explanations for phenomena that are either logically unsound, inconsistent with established theories, or based on fabricated intermediate steps, even if the final conclusion appears plausible. This is particularly dangerous as it can create a false sense of understanding. For instance, an MLLM might "derive" a correct mathematical formula through a series of incorrect algebraic manipulations.
    \item \textbf{Exploratory Hallucination:} This is a more nuanced category. It refers to the generation of novel, plausible-sounding hypotheses or ideas that are not yet validated or directly supported by existing data. While technically a "hallucination" (as it's not grounded in current certainties), this type can be beneficial in the "Creative Hypothesis Generation" stage if properly managed and clearly identified as speculative. The risk lies in presenting such exploratory outputs as established facts.
\end{itemize}

\subsection{Existing Mitigation Strategies} 

Several strategies have been developed to mitigate hallucinations in general MLLM applications, which can be adapted, with limitations, to scientific reasoning:

\begin{itemize}[leftmargin=*]
    \item \textbf{Knowledge Retrieval Augmentation (RAG):} This involves grounding the MLLM's responses by retrieving relevant information from external, trusted knowledge bases (\textit{e.g.,} scientific databases, curated literature). The model then uses this retrieved context to formulate its answer, reducing reliance on its parametric memory which can be a source of hallucinations.
    \item \textbf{Consistency Checks:}
        \begin{itemize}
            \item \textit{Self-consistency:} Generating multiple reasoning paths or answers for the same query and selecting the most consistent or frequently occurring one.
            \item \textit{Cross-modal consistency:} For MLLMs, ensuring that information derived from text aligns with information from images (\textit{e.g.,} a diagram in a physics problem should match the textual description).
            \item \textit{Logical consistency:} Evaluating the logical coherence of the generated reasoning steps.
        \end{itemize}
    \item \textbf{Process-Based Supervision (and Process Reward Models - PRMs):} Instead of only rewarding the correctness of the final answer, these methods involve supervising or rewarding the intermediate steps of the reasoning process. This encourages the model to follow valid inferential pathways, making it less likely to "jump" to conclusions through hallucinated steps.
    \item \textbf{Human Feedback (RLHF/RLAIF):} Reinforcement Learning from Human Feedback (and its variants like RLAIF) involves training the MLLM based on human ratings of its outputs. Humans can penalize hallucinated content, guiding the model towards more factual and reliable responses.
\end{itemize}

\subsection{Why Current Mitigation Methods Fall Short in Scientific Reasoning} 

Despite their utility, current hallucination mitigation strategies face significant hurdles when applied to the rigorous demands of scientific reasoning:

\begin{itemize}[leftmargin=*]
    \item \textbf{Complexity and Nuance of Scientific Knowledge:}
        \begin{itemize}
            \item Scientific knowledge is vast, deeply interconnected, and often highly nuanced or domain-specific. RAG systems might struggle to retrieve the \textit{exact} precise piece of information needed or may misinterpret the context of retrieved documents. The sheer volume can also lead to retrieving conflicting information.
            \item Many scientific concepts are abstract and symbolic (\textit{e.g.,} in quantum mechanics or advanced mathematics), making it difficult for models to truly "understand" and ground them, even with retrieved text.
        \end{itemize}
    \item \textbf{The Need for Exploration and Controlled Creativity:}
        \begin{itemize}
            \item Aggressive hallucination mitigation, designed to enforce strict factuality, can inadvertently stifle the model's ability to engage in exploratory reasoning or generate novel hypotheses (critical for Stage 4). The challenge is to allow for "beneficial" exploratory generation while suppressing detrimental factual or conceptual hallucinations—a balance current methods struggle to achieve.
        \end{itemize}
    \item \textbf{Lack of Granular Explainability and Verifiability:}
        \begin{itemize}
            \item Even if a mitigation technique reduces overt hallucinations, the internal reasoning process of MLLMs often remains a black box. For scientific applications, it's not enough for an answer to be correct; the reasoning must be transparent, verifiable, and align with established scientific methods. Current methods don't inherently guarantee this level of scrutiny.
            \item Error propagation in multi-step scientific reasoning is a major issue. A subtle, un-caught hallucination in an early step can invalidate the entire chain, and PRMs might not be sufficiently fine-grained to catch all such nuanced errors in complex scientific domains.
        \end{itemize}
    \item \textbf{Data Limitations for Scientific Domains:}
        \begin{itemize}
            \item High-quality, large-scale, and \textit{verified} multimodal scientific datasets specifically curated for training MLLMs to avoid domain-specific hallucinations are scarce.
            \item Training effective PRMs or RLAIF systems for science requires extensive, expert-annotated data on correct reasoning processes, which is costly and time-consuming to produce across diverse scientific fields.
            \item Models trained on existing scientific literature might inherit biases or outdated information present in that corpus, leading to "hallucinations" relative to the current state of knowledge.
        \end{itemize}
\end{itemize}

In conclusion, while existing mitigation techniques provide a starting point, the unique demands of scientific reasoning—its need for precision, depth, verifiability, and controlled creativity—necessitate the development of more sophisticated, domain-aware, and interpretable approaches to manage and understand hallucinations in MLLMs.

\section{Multimodal Scientific (M)LLMs}
\label{app:multimodal_scillm}
Scientific (M)LLMs represent an emerging class of models that integrate multiple modalities to tackle complex problems across various scientific disciplines, as summarized in Table \ref{tab:multimodal_scillm}. These models leverage massive datasets combining text, images, graphs, and other forms of scientific data to provide deeper insights and advanced reasoning capabilities. Unlike traditional models, which typically focus on either textual or visual information, scientific MLLMs are designed to harmonize and synthesize diverse sources of scientific knowledge, enabling enhanced performance in tasks such as mathematical problem solving, chemical reaction prediction, biological analysis, and physical simulations \cite{morris2023scientists,pei2024leveraging,reddy2024towards,zhang2024comprehensive}.
\begin{table*}[t]
\caption{Summary of scientific (M)LLMs.}
\label{tab:multimodal_scillm}
\setlength{\tabcolsep}{6pt}
\renewcommand\arraystretch{1.05}
\begin{adjustbox}{max width=\textwidth}
\begin{tabular}{ccccccccc}
\hline
\multirow{2}{*}{\textbf{Paper}} &
\multirow{2}{*}{\makecell{\textbf{Organization}}} &
\multirow{2}{*}{\makecell{\textbf{Venue}}} &
\multirow{2}{*}{\makecell{\textbf{Multimodal}}} &
\multirow{2}{*}{\makecell{\textbf{Parameter}}} &
\multicolumn{4}{c}{\textbf{Domain(s)}} \\
\cline{6-9}
& & & & & Math & Physics & Chemistry & Biology \\
\hline

\textbf{General-Reasoner}~\cite{ma2025generalreasoner} &
Waterloo & NeurIPS'25 &  &
\makecell{4B/7B/14B} &
\checkmark & \checkmark & \checkmark & \checkmark \\

\textbf{AI-Reasoner}~\cite{tang2025airesearcher} &
HKU & NeurIPS'25 &  &
- &
\checkmark & \checkmark & \checkmark & \checkmark \\

\textbf{OmniScience}~\cite{prabhakar2025omniscience} &
SES AI &  &  &
\makecell{70B} &
\checkmark & \checkmark & \checkmark & \checkmark \\

\textbf{Galactica}~\cite{taylor2022galactica} &
Meta &  & \checkmark &
\makecell{125M/1.3B/6.7B/\\30B/120B} &
\checkmark & \checkmark & \checkmark & \checkmark \\

\textbf{LLM-SR}~\cite{Shojaee2024LLMSRSE} &
Virginia Tech &  &  &
\makecell{8x7B} &
\checkmark & \checkmark &  & \checkmark \\

\textbf{SciReasoner}~\cite{wang2025scireasoner} &
Shanghai AI Lab &  &  &
\makecell{1.7B/8B} &
 & \checkmark & \checkmark & \checkmark \\

\textbf{SciLitLLM}~\cite{Li2024SciLitLLMHT} &
USTC &  &  &
\makecell{7B/14B} &
 & \checkmark & \checkmark & \checkmark \\

\textbf{Darwin series}~\cite{Xie2023DARWINSD} &
\makecell{University of\\New South Wales} &  &  &
\makecell{7B} &
 & \checkmark & \checkmark &  \\

\textbf{SPMM}~\cite{SPMM} &
KAIST & \makecell{Nature\\Communications'24} & \checkmark &
-- &  &  & \checkmark & \checkmark \\

\textbf{InstructMol}~\cite{Cao2023InstructMolMI} &
\makecell{IDEA \& HKUST} & COLING'25 & \checkmark &
\makecell{7B} &
 &  & \checkmark & \checkmark \\

\textbf{BioinspiredLLM}~\cite{Luu2023BioinspiredLLMCL} &
MIT & \makecell{Advanced\\Science'24} &  &
\makecell{13B} &
 &  & \checkmark & \checkmark \\

\textbf{nach0}~\cite{Livne2023nach0MN} &
NVIDIA & \makecell{Chemical\\Science'24} & \checkmark &
\makecell{250M/780M} &
 &  & \checkmark & \checkmark \\

\textbf{Mole-BERT}~\cite{Xia2023MoleBERTRP} &
Westlake University & ICLR'23 &  &
-- &  &  & \checkmark & \checkmark \\

\textbf{BioReason}~\cite{fallahpour2025bioreason} &
University of Toronto &  &  &
\makecell{1B/4B} &
 &  &  & \checkmark \\

\textbf{BioGPT}~\cite{Luo2022BioGPTGP} &
Microsoft & \makecell{Briefings in\\Bioinformatics'22} &  &
\makecell{355M} &
 &  &  & \checkmark \\

\textbf{Evolla}~\cite{Zhou2025evolla} &
Westlake University &  & \checkmark &
\makecell{80B} &
 &  &  & \checkmark \\

\textbf{Prollama}~\cite{Lv2024ProLLaMAAP} &
Peking University &  &  &
\makecell{7B} &
 &  &  & \checkmark \\

\textbf{ether0}~\cite{narayanan2025training} &
\makecell{FutureHouse Inc} & NeurIPS'25 &  &
24B &  &  & \checkmark &  \\

\textbf{MOOSE-Chem2}~\cite{yang2025moose} &
\makecell{NTU \& Shanghai AI Lab} & NeurIPS'25 &  &
-- &  &  & \checkmark &  \\

\textbf{Perovskite-LLM}~\cite{liu2025perovskitellm} &
\makecell{HKUST(GZ)} & EMNLP Findings'25 &  &
8B &  &  & \checkmark &  \\

\textbf{Chemdfm}~\cite{Zhao2024ChemDFMAL} &
\makecell{Shanghai Jiaotong\\University} &  &  &
\makecell{13B} &
 &  & \checkmark &  \\

\textbf{ChemDFM-X}~\cite{Zhao2024ChemDFMXTL} &
\makecell{Shanghai Jiaotong\\University} &  & \checkmark &
\makecell{8B} &
 &  & \checkmark &  \\

\textbf{ChemLLM}~\cite{Zhang2024ChemLLMAC} &
Shanghai AI Lab &  &  &
\makecell{7B} &
 &  & \checkmark &  \\

\textbf{GIT-Mol}~\cite{Liu2023GITMolAM} &
Peng Cheng Lab &
\makecell{Computers in Biology\\and Medicine'24} &
\checkmark &
\makecell{700M} &
 &  & \checkmark &  \\

\textbf{MolGPT}~\cite{Bagal2021MolGPTMG} &
\makecell{International Institute of\\Information Technology} &
\makecell{Journal of Chemical\\Information and Modeling'21} &
 & \makecell{6M} &
 &  & \checkmark &  \\

\textbf{MOOSE-Chem}~\cite{Yang2024MOOSEChemLL} &
\makecell{NTU \& Shanghai AI Lab} &  &  &
-- &  &  & \checkmark &  \\

\textbf{BatGPT-Chem}~\cite{Yang2024BatGPTChemAF} &
\makecell{Shanghai Jiaotong\\University} &  &  &
\makecell{15B} &
 &  & \checkmark &  \\

\textbf{DARWIN 1.5}~\cite{Xie2024DARWIN1L} &
\makecell{University of\\New South Wales} &  &  &
\makecell{7B} &
 &  & \checkmark &  \\

\textbf{MolMetaLM}~\cite{Wu2024MolMetaLMAP} &
Central South University &  &  &
-- &  &  & \checkmark &  \\

\textbf{SMI-TED}~\cite{Soares2024SMI-TED} &
IBM &  &  &
\makecell{289M} &
 &  & \checkmark &  \\

 \textbf{MathCoder2}~\cite{lu2024mathcoder2} &
CUHK & ICLR'25 &  &
\makecell{7B} &
\checkmark &  &  &  \\

\textbf{MathCoder-VL}~\cite{wang2025mathcoder} &
CUHK & ACL'25 Findings & \checkmark &
\makecell{2B/8B} &
\checkmark &  &  &  \\

\textbf{MathCoder}~\cite{Wang2023MathCoderSC} &
CUHK & ICLR'24 &  &
\makecell{7B/13B} &
\checkmark &  &  &  \\

\textbf{MAmmoTH1}~\cite{Yue2023MAmmoTHBM} &
UWaterloo & ICLR’24 &  &
\makecell{7B/13B/70B} &
\checkmark &  &  &  \\

\textbf{Math-LLaVA}~\cite{Shi2024MathLLaVABM} &
NUS & \makecell{EMNLP Findings'24} &  &
\makecell{13B} &
\checkmark &  &  &  \\

\textbf{JiuZhang 2.0}~\cite{Zhao2023JiuZhang2.0A} &
\makecell{RUC \& iFLYTEK} & KDD'23 &  &
-- & \checkmark &  &  &  \\

\textbf{JiuZhang 1.0}~\cite{Zhao2022JiuZhang1.0AC} &
\makecell{RUC \& iFLYTEK} & KDD'22 &  &
\makecell{145M} &
\checkmark &  &  &  \\

\textbf{Minerva}~\cite{Lewkowycz2022Minerva} &
Google & NeurIPS'22 &  &
\makecell{8B/62B/540B} &
\checkmark &  &  &  \\

\textbf{Hypertree Proof Search}~\cite{Lample2022HyperTreePS} &
Meta & NeurIPS'22 &  &
-- & \checkmark &  &  &  \\

\textbf{Qwen2.5-Math}~\cite{Yang2024Qwen25MathTR} &
Alibaba &  &  &
\makecell{1.5B/7B/72B} &
\checkmark &  &  &  \\

\textbf{Qwen2-Math}~\cite{yang2024qwen2math} &
Alibaba &  &  &
\makecell{1.5B/7B/72B} &
\checkmark &  &  &  \\

\textbf{Qwen2-Math-Instruct}~\cite{yang2024qwen2math} &
Alibaba &  &  &
\makecell{1.5B/7B/72B} &
\checkmark &  &  &  \\

\textbf{MathGPT}~\cite{mathgpt} &
TAL Group &  & \checkmark &
\makecell{130B} &
\checkmark &  &  &  \\

\textbf{math-specialized Gemini 1.5 Pro}~\cite{team2024geminipro} &
Google &  & \checkmark &
-- & \checkmark &  &  &  \\

\textbf{InternLM2-Math}~\cite{Ying2024InternLMMathOM} &
Shanghai AI Lab &  &  &
\makecell{1.8B/7B/20B/\\8x22B} &
\checkmark &  &  &  \\

\textbf{InternLM2.5-StepProver}~\cite{Wu2024InternLM25StepProverAA} &
Shanghai AI Lab &  &  &
\makecell{7B} &
\checkmark &  &  &  \\

\textbf{Llemma}~\cite{Azerbayev2023LlemmaAO} &
\makecell{Princeton University\\\& Eleuther AI} &  &  &
\makecell{7B/34B} &
\checkmark &  &  &  \\

\textbf{ChatGLM-Math}~\cite{Xu2024ChatGLMMathIM} &
Zhipu AI &  &  &
\makecell{32B} &
\checkmark &  &  &  \\

\textbf{MetaMath}~\cite{Yu2023MetaMathBY} &
\makecell{Cambridge \& Huawei} &  &  &
\makecell{7B/13B/70B} &
\checkmark &  &  &  \\

\textbf{MathGLM}~\cite{yang2023mathglm} &
\makecell{Tsinghua \& Zhipu AI} &  &  &
\makecell{10M/100M/335M/\\500M/2B/6B/10B} &
\checkmark &  &  &  \\

\textbf{MathGLM-Vision}~\cite{Yang2024MathGLMVisionSM} &
\makecell{Tsinghua \& Zhipu AI} &  & \checkmark &
\makecell{9B/19B/32B} &
\checkmark &  &  &  \\

\textbf{Skywork-13B-Math}~\cite{skyworkmath} &
SkyworkAI &  & \checkmark &
\makecell{7B/13B} &
\checkmark &  &  &  \\

\textbf{DeepSeekMath-V2}~\cite{shao2025deepseekmathv2} &
DeepSeek AI &  &  &
\makecell{685B} &
\checkmark &  &  &  \\

\textbf{DeepSeekMath}~\cite{Shao2024DeepSeekMathPT} &
DeepSeek AI &  &  &
\makecell{7B} &
\checkmark &  &  &  \\

\textbf{DeepSeekProver-V1}~\cite{Xin2024DeepSeekProverAT} &
DeepSeek AI &  &  &
\makecell{7B} &
\checkmark &  &  &  \\

\textbf{DeepSeek-Prover-V.15}~\cite{xin2024deepseekproverv15harnessingproofassistant} &
DeepSeek AI &  &  &
\makecell{7B} &
\checkmark &  &  &  \\

\textbf{Mathstral}~\cite{mistral2024mathstral} &
Mistral AI &  &  &
\makecell{7B} &
\checkmark &  &  &  \\

\textbf{JiuZhang 3.0}~\cite{Zhou2024JiuZhang30EI} &
\makecell{RUC \& iFLYTEK} &  &  &
\makecell{7B/8B} &
\checkmark &  &  &  \\

\textbf{Math-LLM}~\cite{liu2024mathllm} &
\makecell{East China Normal\\University} &  & \checkmark &
\makecell{8.26B/7B/72B} &
\checkmark &  &  &  \\

\textbf{GPT-f}~\cite{Polu2020gpt-f} &
OpenAI &  &  &
\makecell{160M/400M/700M/} &
\checkmark &  &  &  \\

\textbf{Rho-Math}~\cite{lin2024rhomath} &
Microsoft &  &  &
\makecell{1B/7B} &
\checkmark &  &  &  \\

\textbf{WizardMath}~\cite{Luo2023WizardMathEM} &
Microsoft &  &  &
\makecell{7B/70B} &
\checkmark &  &  &  \\

\textbf{Xwin-LM}~\cite{Ni2024XwinLMSA} &
Microsoft &  &  &
\makecell{7B/13B/70B} &
\checkmark &  &  &  \\

\textbf{MAmmoTH2}~\cite{Yue2024MAmmoTH2SI} &
UWaterloo &  &  &
\makecell{7B/8B} &
\checkmark &  &  &  \\

\textbf{GAIRMath-Abel}~\cite{gairmathabel} &
\makecell{Shanghai Jiaotong\\University} &  &  &
\makecell{7B/13B/70B} &
\checkmark &  &  &  \\

\textbf{KwaiYiiMath}~\cite{Fu2023KwaiYiiMathTR} &
Kuaishou &  &  &
\makecell{13B} &
\checkmark &  &  &  \\

\textbf{k0-math}~\cite{k0math} &
Moonshot AI &  &  &
-- & \checkmark &  &  &  \\

\textbf{NuminaMath}~\cite{numina_math_7b} &
Numina &  &  &
\makecell{7B/72B} &
\checkmark &  &  &  \\
\hline
\end{tabular}
\end{adjustbox}
\vskip-0.5em
\end{table*}

\section{Discussion of More Modalities in Scientific Reasoning}
\label{app:more_modality}

While current MLLMs have demonstrated remarkable progress, their sensory input is predominantly confined to text and static images. This limitation curtails their ability to fully comprehend and reason about the multifaceted nature of scientific phenomena, which often unfold dynamically, possess intricate three-dimensional structures, or generate non-visual sensory data. To significantly advance scientific reasoning, MLLMs must evolve to integrate a richer spectrum of modalities, including audio, video, 3D structural data, and other sensor-derived information. This expansion is not merely about processing more data types but about enabling a more holistic and nuanced understanding essential for complex scientific inquiry.

\subsection{Current Limitations and the Imperative for Richer Modalities}
The current text-image paradigm in MLLMs, while powerful, presents several inherent limitations when applied to the diverse needs of scientific reasoning:
\begin{itemize}[leftmargin=*]
    \item \textbf{Restricted Modality Spectrum:} Most MLLMs excel with visual and textual data but struggle to incorporate or meaningfully reason over audio (\textit{e.g.,} sonification of data, acoustic signatures of experiments), video (\textit{e.g.,} dynamic processes, experimental procedures), and 3D data (\textit{e.g.,} molecular configurations, geological strata). This restricts their ability to capture the full context of many scientific observations.
    \item \textbf{Inadequate for Dynamic Phenomena:} Scientific processes are often dynamic. For instance, observing a chemical reaction, the growth of a biological culture, or the time-evolution of a physical system requires processing temporal information embedded in video or sequential sensor readings. Static images and text alone cannot capture these crucial dynamic aspects.
    \item \textbf{Insufficient for Spatial Complexity:} Many scientific domains, such as chemistry (molecular structures), biology (protein folding, anatomical systems), materials science (crystal lattices), and earth sciences (geophysical models), fundamentally rely on understanding complex 3D spatial relationships. Current MLLMs often lack the sophisticated 3D geometric reasoning capabilities required.
    \item \textbf{Limited Cross-Modal Association for Complex Reasoning:} Scientific reasoning frequently involves drawing inferences across multiple, diverse data streams. For example, correlating a change in a visual indicator (image/video) with a specific sound (audio) and a corresponding data log (text/tabular) to deduce a causal link in an experiment demands finer-grained and more diverse modality interactions than currently supported.
    \item \textbf{Domain-Specific Knowledge Integration Challenges:} Effectively interpreting specialized modalities (\textit{e.g.,} spectrographic data, sonified outputs) necessitates not just raw data processing but also deep integration with domain-specific ontologies and knowledge bases to contextualize the sensory input accurately.
\end{itemize}

\subsection{Technical Approaches for Broader Modality Integration}
Addressing these limitations requires developing robust technical approaches for encoding and integrating a wider array of modalities, tailored to their unique characteristics and scientific relevance:
\begin{itemize}[leftmargin=*]
    \item \textbf{Audio Processing:}
    \begin{itemize}
        \item \textit{Feature Extraction:} Leveraging established techniques like spectrograms or Mel-Frequency Cepstral Coefficients (MFCCs) to convert raw audio signals from experiments (\textit{e.g.,} equipment sounds, material stress responses) or sonified data into feature representations suitable for neural networks.
        \item \textit{Model Adaptation:} Developing specialized audio encoders or adapting existing ones, and aligning their latent spaces with those of text and image encoders through cross-modal attention mechanisms or joint embedding strategies. This allows, for instance, an MLLM to associate the sound of a failing component with its visual depiction and textual description of failure modes.
    \end{itemize}
    \item \textbf{Video Understanding:}
    \begin{itemize}
        \item \textit{Spatiotemporal Modeling:} Employing 3D Convolutional Neural Networks (3D CNNs), Video Transformers, or other video-language models to capture both spatial features within frames and temporal dynamics across frames. This is critical for analyzing experimental procedures, observing cellular motility, or tracking particle trajectories.
        \item \textit{Keyframe and Event Detection:} Implementing mechanisms to automatically identify and focus on critical events or keyframes within a scientific video (\textit{e.g.,} phase transition, initiation of a reaction), thereby reducing computational load and highlighting salient information for reasoning.
    \end{itemize}
    \item \textbf{3D Structural Data Processing:}
    \begin{itemize}
        \item \textit{Geometric Encoding:} Utilizing encoders like PointNet, Point Transformer, or Graph Neural Networks (GNNs) to process irregular 3D data such as point clouds (\textit{e.g.,} from LiDAR scans of geological sites) or molecular meshes.
        \item \textit{Incorporating Physical and Chemical Constraints:} Enhancing structural understanding by integrating domain-specific knowledge, such as bond lengths and angles in molecular modeling or material properties in engineering design, directly into the encoding or reasoning process. This ensures that interpretations are scientifically plausible.
    \end{itemize}
    \item \textbf{Sensor and Time-Series Data:}
    \begin{itemize}
        \item \textit{Specialized Encoders:} Using Recurrent Neural Networks (RNNs), 1D CNNs, or Transformers adapted for time-series data from various sensors (\textit{e.g.,} temperature, pressure, EEG, spectroscopy).
        \item \textit{Temporal Alignment:} Developing methods to align asynchronous and heterogeneous sensor streams with other modalities like video or textual logs of experimental events.
    \end{itemize}
\end{itemize}

\subsection{Advanced Cross-Modal Interaction and Scientific Reasoning}
The true power of incorporating more modalities lies in enabling more sophisticated cross-modal interactions and reasoning capabilities:
\begin{itemize}[leftmargin=*]
    \item \textbf{Unified Multimodal Alignment:}
    \begin{itemize}
        \item \textit{Shared Embedding Space:} Designing a common embedding space, potentially through CLIP-style contrastive learning extended to audio, video, and 3D data, where features from diverse modalities representing the same scientific concept are brought closer together.
        \item \textit{Dynamic Modality Weighting:} Implementing mechanisms that allow the model to dynamically assign importance (weights) to different modalities based on the specific scientific task, context, or data quality, ensuring robustness and flexibility.
    \end{itemize}
    \item \textbf{Enhanced Scientific Reasoning Modules:}
    \begin{itemize}
        \item \textit{Multimodal Causal Graph Generation:} Constructing more comprehensive causal relationship graphs by leveraging evidence from diverse inputs. For example, observing a video of an experimental setup, listening to audio cues indicating a process stage, and reading instrument outputs (text/tabular) can collectively inform the nodes and edges of a causal graph explaining an observed phenomenon (\textit{e.g.,} "increased gas pressure [sensor] caused by heating element activation [video/log] led to audible valve release [audio]").
        \item \textit{Generative Reasoning with Richer Evidence:} Empowering language models to generate more nuanced and well-supported scientific explanations, hypotheses, or experimental designs by integrating and citing evidence from audio logs, video segments, 3D visualizations, and sensor data alongside textual knowledge.
    \end{itemize}
\end{itemize}

\subsection{Task-Driven Training Strategies and Future Directions}
Realizing the potential of MLLMs with expanded modalities requires innovative training strategies and a forward-looking research agenda:
\begin{itemize}[leftmargin=*]
    \item \textbf{Multi-Task Learning for Comprehensive Understanding:}
    \begin{itemize}
        \item Training models on a combination of tasks, including fundamental modality understanding (\textit{e.g.,} audio event classification, video action recognition, 3D object recognition) alongside higher-level scientific reasoning tasks (\textit{e.g.,} predicting experimental outcomes, inferring material properties from multimodal sensor data).
    \end{itemize}
    \item \textbf{Leveraging Domain-Specific Knowledge Graphs:}
    \begin{itemize}
        \item Incorporating expert-annotated scientific knowledge graphs (ontologies, relational databases) as prior constraints or auxiliary information during training. This can guide the model in interpreting complex multimodal inputs within the correct scientific framework (\textit{e.g.,} linking a 3D protein structure to its known functions and interaction pathways).
    \end{itemize}
    \item \textbf{Addressing Data Scarcity:}
    \begin{itemize}
        \item Developing techniques for data augmentation, synthetic data generation (\textit{e.g.,} simulating experimental videos or 3D molecular interactions), and leveraging weakly supervised or self-supervised learning methods, given that large-scale, richly annotated multimodal scientific datasets are often scarce.
    \end{itemize}
    \item \textbf{Future Outlook:}
    The path forward involves creating benchmark datasets that encompass these richer modalities within scientific contexts. Furthermore, research should explore how these expanded sensory capabilities can enable truly interactive MLLMs that can actively participate in the scientific process, for instance, by suggesting modifications to an experimental setup based on real-time video and sensor feedback. Such advancements will be pivotal in unlocking new frontiers in AI-assisted scientific discovery.
\end{itemize}

\section{Discussion of Alternative Views}
\label{app:alternative_details}

While this paper champions the significant potential of MLLMs to advance scientific reasoning, it is crucial to engage with pertinent alternative viewpoints and concerns that challenge this optimism. Acknowledging these perspectives allows for a more nuanced understanding of the MLLM adoption pathway in science.

\subsection{Domain-Specific Models as a Superior Alternative}
\label{sec:alternative1}

\begin{figure*}[t!]
    \centering
    \includegraphics[width=1 \linewidth]{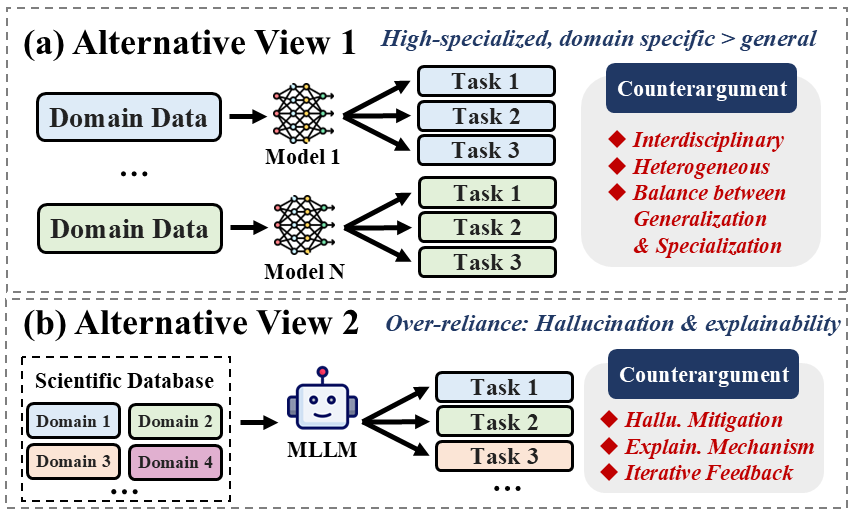}
    \caption{Illustrations of alternative view 1 (a) and view 2 (b), as well as our corresponding counterarguments.}
    \label{fig:alternative_views}
\end{figure*}

One argument suggests that \textit{highly specialized, domain-specific models tailored to individual scientific disciplines may outperform general-purpose MLLMs in reasoning tasks}, as indicated in Figure \ref{fig:alternative_views} (a). Proponents of this view highlight that scientific reasoning often requires deep domain expertise, nuanced understanding, and customized data processing pipelines that are difficult to replicate in generalized multimodal architectures \cite{zhang2024scientific,barman2025large}. For example, domain-specific models like AlphaFold for protein structure prediction \cite{jumper2021highly} or symbolic computation tools for solving mathematical problems \cite{mirzadeh2024gsm,lu2021inter} excel precisely because of their narrow focus.

\textbf{Counterargument.} While domain-specific models have demonstrated remarkable success in narrow applications, they lack the flexibility to generalize across multiple domains or integrate diverse modalities of data. Scientific reasoning increasingly involves interdisciplinary approaches, such as bioinformatics combining biology and computational techniques, or climate science requiring both geospatial and textual analysis \cite{reddy2024towards}. MLLMs, by design, offer the ability to process and reason over heterogeneous data sources, enabling a broader and more integrative approach \cite{zhang2024comprehensive}. Furthermore, domain-specific knowledge can still be fine-tuned within MLLMs, allowing these systems to leverage the best of both generalization and specialization \cite{chen2023fine}.

\textbf{More Elaboration} can be seen as follows:
\begin{itemize}[leftmargin=*]
    \item \textbf{Necessity of Customized Architectures and Data Handling:}
    \begin{itemize}
        \item \textit{What creates the necessity of a customized model?} Many scientific domains rely on unique, highly structured data formats (\textit{e.g.,} SMILES strings in chemistry, FASTA sequences in genomics, PDB files for protein structures, specific notations in physics diagrams or mathematical proofs). General MLLMs, even with multimodal capabilities, may lack the inherent inductive biases or fine-grained understanding to parse, interpret, and reason over these specialized representations with the required precision. A custom model can be built from the ground up to "speak" the native language of that data.
        \item Domain-specific models can embed established physical laws, chemical reaction rules, or biological pathways more explicitly, either through architectural design or by training on curated datasets that heavily emphasize these principles. This contrasts with MLLMs that must infer these rules implicitly from broader, potentially noisier data.
        \item The sheer precision required for tasks like predicting quantum mechanical properties, drug-target interaction affinities, or complex engineering tolerances might necessitate models whose entire objective function and training data are geared towards minimizing error in that specific, narrow domain, rather than general plausibility.
    \end{itemize}

    \item \textbf{Knowledge Depth and Nuance:}
    \begin{itemize}
        \item Scientific breakthroughs often require profound depth of knowledge and nuanced understanding within a very specific sub-field. While MLLMs aim for breadth, they might only achieve a superficial understanding in areas requiring years of human expert training. The "curse of dimensionality" applies to knowledge as well; covering all scientific domains deeply in one model is an immense challenge.
        \item Training datasets for general MLLMs, even if including scientific papers, might be diluted by vast amounts of non-scientific text, or may not capture the long tail of niche, but critical, scientific knowledge that specialized databases and models incorporate.
    \end{itemize}
\end{itemize}
While the counterargument in the main paper highlights MLLM flexibility for interdisciplinary tasks, proponents of specialization argue that many critical scientific problems are bottlenecked by depth in a single discipline, not breadth. They contend that fine-tuning an MLLM might offer some domain adaptation, but it's unlikely to match a model purpose-built for the intricacies of that domain.
\subsection{Risks of Over-reliance on MLLMs}
\label{sec:alternative2}

Another valid concern is \textit{the potential over-reliance on MLLMs, which could exacerbate issues such as hallucination and lack of explainability}, as shown in Figure \ref{fig:alternative_views} (b). Critics argue that MLLMs, while powerful, are prone to generating plausible-sounding but incorrect or unsubstantiated outputs. This risk is particularly concerning in scientific reasoning, where accuracy and rigor are paramount \cite{bai2024hallucination}. Additionally, the black-box nature of these models makes it difficult for researchers to validate their reasoning processes or trust their conclusions, which could hinder their adoption in critical scientific applications \cite{cambria2024xai,rodis2024multimodal,dang2024explainable}.

\textbf{Counterargument.} These concerns are valid and underscore the need for a cautious and measured approach to MLLM adoption. However, rather than dismissing MLLMs outright, these issues highlight areas for improvement. For example, integrating explainability mechanisms, such as visual attention maps \cite{chefer2021generic,dehimi2024attention} or rationale generation \cite{hu2024learning,wu2024usable}, can enhance transparency. Additionally, hybrid models that combine MLLMs with symbolic reasoning \cite{li2024deceptive,zhou2024mitigating} or expert systems \cite{guan2024mitigating,niu2024mitigating} can mitigate risks while maintaining strengths of multimodal reasoning. Finally, iterative feedback loops and human-in-the-loop systems can ensure reliability in reasoning workflows \cite{xiao2024detecting,zou2024look,zheng2024reefknot}.

\textbf{More Elaboration} can be seen as follows:
\begin{itemize}[leftmargin=*]
    \item \textbf{The Amplified Peril of Scientific Hallucination:}
    \begin{itemize}
        \item Hallucinations in a general chatbot might be an annoyance; in science, they can be dangerous or resource-wasting. This isn't just about generating factually incorrect statements, but also subtle misinterpretations of complex multimodal inputs (\textit{e.g.,} misreading a crucial detail in a micrograph or circuit diagram) that lead to flawed reasoning chains.
        \item The "plausible-sounding but incorrect" nature of MLLM outputs is especially insidious in science, where a subtly flawed equation, a non-existent chemical reaction pathway, or an incorrectly inferred biological interaction could derail research or lead to unsafe experimental protocols.
    \end{itemize}

    \item \textbf{The Imperative of Explainability and Verifiability:}
    \begin{itemize}
        \item The scientific method fundamentally relies on transparency, reproducibility, and the ability to scrutinize the reasoning process. The "black-box" nature of many MLLMs directly conflicts with this. How does one verify the multi-step reasoning an MLLM used to arrive at a novel hypothesis from a complex set of inputs (\textit{e.g.,} a research paper, experimental data, and a diagram)?
        \item Current explainability techniques (\textit{e.g.,} attention maps, rationale generation) often provide superficial or post-hoc justifications that may not reflect the true internal "reasoning" of the model, if such a coherent process even exists in a human-understandable form. This makes debugging errors or building trust in novel, MLLM-generated insights exceptionally difficult.
    \end{itemize}

    \item \textbf{Bias Propagation and Deskilling Concerns:}
    \begin{itemize}
        \item MLLMs trained on existing scientific literature and datasets will inevitably inherit and potentially amplify existing biases within those sources (\textit{e.g.,} overrepresentation of certain research topics, demographic groups in clinical data, or established theories at the expense of emerging ones).
        \item A more philosophical concern is the potential for "deskilling." If researchers become overly reliant on MLLMs for hypothesis generation, data interpretation, or even experimental design, there's a risk of atrophying fundamental human scientific reasoning and critical thinking skills, reducing scientists to sophisticated prompt engineers.
    \end{itemize}
\end{itemize}
The main paper's counterargument focuses on mitigation strategies. However, critics would argue that the current state of these mitigations is insufficient for high-stakes scientific applications, and the burden of proof lies in demonstrating consistent reliability, deep interpretability, and bias control before widespread adoption in critical scientific endeavors. The very nature of MLLMs---learning statistical patterns rather than causal mechanisms---might pose a fundamental limit to their trustworthiness in a domain that prizes causal understanding.

\section{Clarification of LLM Usage}
\label{app:llm_usage}
In the preparation of this manuscript, the authors leveraged Gemini-Pro-2.5 as an assistive writing tool. The use of the LLM was strictly confined to an editorial capacity, primarily for refining language, improving grammatical correctness, and enhancing the clarity and readability of the text. The core intellectual contributions—including the central thesis, the proposed four-stage research roadmap, the analysis of challenges, and the overall structure of the paper—were conceived and developed exclusively by the human authors. The authors meticulously reviewed, edited, and approved all final wording to ensure it accurately reflects their original ideas and arguments, and bear full responsibility for the accuracy and integrity of the content.

\end{document}